\documentclass[a4paper]{article}
\usepackage{arxiv}
\usepackage{times}
\usepackage{hyperref}
\usepackage{multirow}
\usepackage{graphicx}
\usepackage{amsmath,amssymb,amsbsy,amsfonts,amsthm}
\usepackage{algorithm}
\usepackage{algorithmic}
\makeatletter
\renewcommand*{\eqref}[1]{%
  \hyperref[{#1}]{\textup{\tagform@{\ref*{#1}}}}%
}
\makeatother
\usepackage[utf8]{inputenc}
\usepackage{biblatex}
\AtEveryBibitem{
	\clearlist{language}
	\clearfield{doi}
	\clearfield{issn}
	\clearfield{isbn}
	\clearfield{urlyear}
	\ifentrytype{misc}{}{\clearfield{url}}
}
\usepackage{subcaption}
\usepackage{url}
\usepackage{colortbl}
\newcommand{\task}[1]{\mathcal{T}_{#1}}
\renewcommand{\algorithmiccomment}[1]{\bgroup\hfill \small//~#1\egroup}
\title{On the Generalization of Neural Combinatorial Optimization Heuristics}
\renewcommand{\shorttitle}{Generalization of NCO heuristics}
\renewcommand{\headeright}{}
\renewcommand{\undertitle}{}
\author{Sahil Manchanda, Sofia Michel, Darko Drakulic and Jean-Marc Andreoli\\NAVER LABS Europe, Grenoble, France\\
{\small\tt http://www.europe.naverlabs.com}}
\date{April 2022}
\begin{document}

\maketitle

\begin{abstract}
Neural Combinatorial Optimization approaches have recently leveraged the expressiveness and flexibility of deep neural networks to learn efficient heuristics for hard Combinatorial Optimization (CO) problems.
However, most of the current methods lack generalization: for a given CO problem, heuristics which are trained on instances with certain characteristics underperform when tested on instances with different characteristics.
While some previous works have focused on varying the training instances properties, we postulate that a one-size-fit-all model is out of reach.
Instead, we formalize solving a CO problem over a given instance distribution as a separate learning task and investigate meta-learning techniques to learn a model on a variety of tasks, in order to optimize its capacity to adapt to new tasks.
Through extensive experiments, on two CO problems, using both synthetic and realistic instances, we show that our proposed meta-learning approach significantly improves the generalization of two state-of-the-art models.
\end{abstract}
%
\section{Introduction}
Combinatorial optimization (CO) aims at finding optimal decisions within finite sets of possible decisions; the sets being typically so large that exhaustive search is not an option \cite{cook_combinatorial_1997}.
CO problems appear in a wide range of applications such as logistics, transportation, finance, energy, manufacturing, etc.
CO heuristics are efficient algorithms that can compute high-quality solutions but without optimality guarantees.
Heuristics are crucial to CO, not only for applications where optimality is not required, but also for exact solvers, which generally exploit numerous heuristics to guide and accelerate their search procedure~\cite{fischetti_heuristics_2011}.
However, the design of such heuristics heavily relies on problem-specific knowledge, or at least experience with similar problems, in order to adapt generic methods to the setting at hand. This design skill that human experts acquire with experience and that is difficult to capture formally, is a typical signal for which statistical methods may help.
In effect, machine learning has been successfully applied to solve CO problems, as shown in the surveys~\cite{cappart_combinatorial_2021,bengio_machine_2021}.
In particular, \emph{Neural Combinatorial Optimization} (NCO) has shown remarkable results by leveraging the full power and expressiveness of deep neural networks to model and automatically derive efficient CO heuristics.
Among the approaches to NCO, supervised learning~\cite{vinyals_pointer_2015,li_combinatorial_2018,joshi_efficient_2019} and reinforcement learning~\cite{bello_neural_2017,nazari_reinforcement_2018,kool_attention_2019} are the main paradigms.

Despite the promising results of end-to-end heuristic learning, a major limitation of these approaches is their lack of generalization to out-of-training-distribution instances for a given CO problem~\cite{bengio_machine_2021,joshi_learning_2020}. For example, models are generally trained on graphs of a fixed size and perform well on unseen “similar” graphs of the same size. However, when tested on smaller or larger ones, performance degrades drastically. Although size variation is the most reported case of poor generalization, in our study we will show that instances of the same size may still vary enough to cause generalization issues. This limitation might hinder the application of NCO to real-life scenarios where the precise target distribution is often not known in advance and can vary with time. A natural way to alleviate the generalization issue is to train on instances with diverse characteristics, such as various graph sizes~\cite{joshi_learning_2019,lisicki_evaluating_2020,li_learning_2021}. Intuitively this amounts to augmenting the training distribution to make it more likely to correctly represent the target instances.

In this paper, we postulate that a one-size-fit-all model is out of reach.
Instead, we believe that one of the strengths of end-to-end heuristic learning is precisely their adaptation to specific data and the exploitation of the underlying structure to obtain an effective specialized heuristic. Therefore we propose to use instance characteristics to define distributions and consider solving a CO problem over a given {\textit instance distribution} as a separate learning task. We will assume a prior over the target task, by assuming it is part of a given {\textit task distribution}, from which we will sample the training tasks. Note that this is a weaker assumption than most current NCO methods that (implicitly) assume knowing the target distribution at training. At the other extreme, without any assumption on the target distribution, the No Free Lunch Theorems of Machine Learning \cite{wolpert_no_1997} tell us that we cannot expect to do better than a random policy.
In this context, meta-learning~\cite{schmidhuber_evolutionary_1987,ravi_optimization_2017} is a natural approach to obtain a model able to adapt to new unseen tasks. Given a distribution of tasks, the idea of meta-learning is to train a model using a sample of those tasks while optimizing its ability to adapt to each of them. Then at test time, when presented with an unseen task from the same distribution, the model needs to be only fine-tuned using a small amount of data from that task.

\noindent \\
\textbf{Contributions:} We focus on two representative state-of-the-art NCO approaches: (i) the reinforcement learning-based method of~\cite{kool_attention_2019} and the supervised learning approach of~\cite{joshi_efficient_2019}.
In terms of CO problems, we use the well-studied Traveling Salesman Problem (TSP) and Capacitated Vehicle Routing Problem (CVRP).
We first analyze the NCO models' generalization capacity along different instance parameters such as the graph size, the vehicle capacity and the spatial distribution of the nodes and highlight the significant drop in performance on out-of-distribution instances (Section~\ref{sec:genprop}). Then we introduce a model-agnostic meta-learning procedure for NCO, inspired by the first-order meta-learning framework of~\cite{nichol_first-order_2018} and adapt it to both the reinforcement and supervised learning-based NCO approaches (Section~\ref{sec:meta}).
Finally, we design an extensive set of experiments to evaluate the performance of the meta-trained models with different pairs of training and test distributions.
Our contributions are summarized as follows:
\begin{itemize}
    \item \textbf{Problem formalization:} We give the first formalization of the NCO out-of-distribution generalization problem and provide experimental evidence of its impact on two state-of-the-art NCO approaches.
    \item \textbf{Meta-learning framework:} We propose to apply a generic meta-training procedure to learn robust NCO heuristics, applicable to both reinforcement and supervised learning frameworks. To the best of our knowledge we are the first to propose meta-learning in this context and prove its effectiveness through extensive experiments.

    \item \textbf{Experimental evaluation:} We demonstrate experimentally that our proposed meta-learning approach does alleviate the generalization issue. The meta-trained models show a better zero-shot generalization performance than the commonly used multi-task training strategy. In addition, using a  limited number of instances from a new distribution, the fine-tuned meta-NCO models are able to catch-up, and even frequently outperform, the reference NCO models, that were specifically trained on the target distribution. We provide results both on synthetic datasets and the well-established realistic Operations Research datasets \textit{TSPlib} and \textit{CVRPlib}.
    \item \textbf{Benchmarking datasets:} Finally, by extending commonly used datasets, we provide an extensive benchmark of labeled TSP and CVRP instances with a  diverse set of distributions, that we hope will help better evaluate the generalization capability of NCO methods on these problems.
\end{itemize}

\section{Related work}
Several papers have noted the lack of out-of-training-distribution  generalization of current NCO heuristics, e.g.~\cite{bengio_machine_2021,cappart_combinatorial_2021}. In particular, \cite{joshi_learning_2020} explored the role of certain architecture choices and inductive biases of NCO models in their ability to generalize to large-scale TSP problems.  In~\cite{lisicki_evaluating_2020}, the authors proposed a curriculum learning approach to train the attention model of~\cite{kool_attention_2019}, assuming good-quality solutions can be accessed during training and using the corresponding optimality gap to guide the scheduling of training instances of various sizes. The proposed curriculum learning in a semi-supervised setting helped improve the original model's generalization on size. Recently, \cite{fu_generalize_2020} proposed a method  able to generalize to large-scale TSP graphs by combining the predictions of a learned model on small subgraphs and using these predictions to guide a Monte Carlo Tree Search, successfully generalizing to instances with up to 10,000 nodes. Note that both \cite{fu_generalize_2020} and \cite{lisicki_evaluating_2020} are specifically designed to deal with size variation.

One can note that hybrid approaches combining learned components and classical CO algorithms tend to generalize better than end-to-end ones. For example, the learning-augmented local search heuristic of \cite{li_learning_2021} was able to train on relatively small CVRP instances and generalize to instances with up to 3000 nodes. Also recent learned heuristics within branch and bound solvers show a strong generalization ability \cite{nair_learning_2018,zarpellon_parameterizing_2021}. Other approaches that generalize well are based on algorithmic learning. For instance, \cite{georgiev_neural_2020} learns to imitate the Ford-Fulkerson algorithm for maximum bipartite matching, by neural execution of a Graph Neural Network, similar to \cite{velickovic_neural_2019} for other graph algorithms. These methods achieve a strong generalization to larger graphs but at the expense of precisely imitating the steps of existing algorithms.

In this paper we focus on the generalization of end-to-end NCO heuristics. In contrast to previous approaches, we propose a general framework, applicable to both supervised and reinforcement (unsupervised) learning-based NCO methods, and that accounts for any kind of distribution shift, including but not restricted to graph size. To the best of our knowledge, we are the first to propose meta-learning as a generic approach to improve the generalization of any NCO model.

\section{Generalization properties}
\label{sec:genprop}
To analyze the generalization properties of different NCO approaches, we focus on two wide-spread CO problems: (i) the Euclidean Traveling Salesman Problem (TSP), where given a set of nodes in a Euclidean space (typically the plane), the goal is to find a tour of minimal length that visits each node exactly once; and (ii) the Capacitated Vehicle Routing Problem (CVRP), where given a depot node, a set of customer nodes with an associated demand and a vehicle capacity, the goal is to compute a set of routes of minimal total length, starting and ending at the depot, such that each customer node is visited and the sum of demands of customers in each route does not exceed the vehicle capacity. Note that the TSP can be viewed as a special case of the CVRP where the vehicle capacity is infinite.
\subsection{Instance distributions as tasks}
\label{sec:graph-distrib}
\begin{figure}
\centering
\includegraphics[scale=0.29]{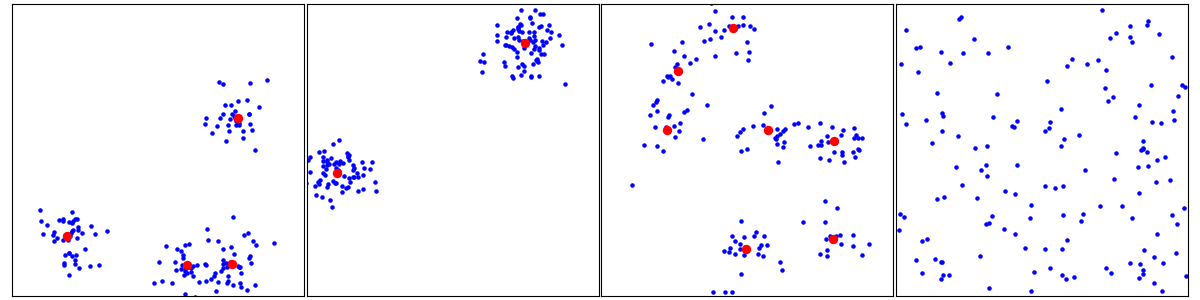}
\caption{\label{fig:task-samples}A sample from each of 4 tasks $\task{N=150,L=1,M}$ (blue points) with $M{=}4,2,7,0$, respectively, from left to right. The red dots are the generated modes.}
\end{figure}

To explore the effect of variability in the training datasets, we consider a specific family $\task{N,M,C,L}$ of instance distributions (tasks), indexed by the following parameters: the {\textit graph size} $N$, the {\textit number of modes} $M$, the {\textit vehicle capacity} $C$ and the {\textit scale} $L$.
Given these parameters, an instance is generated by the following process. When $M{\not=}0$: first, $M$ points, called the modes, are independently sampled by an ad-hoc process which tends to spread them evenly in the unit square; then $N$ points are independently sampled from a balanced mixture of $M$ Gaussian components centered at the $M$ modes, sharing the same diagonal covariance matrix, meant to keep the generated points within relatively small clusters around the modes; finally, the node coordinates are rescaled by a factor $L$. When $M{=}0$: the $N$ points are instead directly sampled uniformly in the unit square then rescaled by $L$. Additionally, in the case of the CVRP problem, the depot is chosen randomly, the vehicle capacity is fixed to $C$ and customer demands are generated as in~\cite{nazari_reinforcement_2018}. Examples of spatial node distributions for various TSP tasks are displayed in Figure~\ref{fig:task-samples}.

\subsection{Measuring the impact of generalization on performance}
To measure the performance of different algorithms on a given task, we sample a set of test instances from that task and apply each algorithm to each of these instances. Since the average length of the resulting tours is biased towards longer lengths, we measure instead the average ``\textit{gap}'' with respect to reference tours. For the TSP, reference is provided by the Concorde solver~\cite{applegate_traveling_2011}, which is exact, so what we report is the true optimality gap; for the CVRP, we use the solutions computed by the state-of-the-art LKH heuristic solver~\cite{helsgaun_extension_2017}, which returns high-quality solutions at the considered instance sizes (near optimality).

We measure the performance (gap) deterioration on generalization of the reinforcement learning based Attention Model of~\cite{kool_attention_2019}, subsequently abbreviated as AM, and the supervised Graph Convolutional Network model of~\cite{joshi_efficient_2019}, subsequently abbreviated as GCN.
We consider several classes of tasks of the form $\task{N,M,C,L}$ obtained by varying, in each class, only one of the parameters\footnote{Except with CVRP where, as in previous work~\cite{nazari_reinforcement_2018}, changes to $C$ and $N$ are coupled.} $N,M,C,L$. For each class and each task in that class, we train each model on that task only and test it on each of the tasks in the same class, thus including the training one.
The main results for the AM model are reported in Table ~\ref{gen:AM_optgap}.

As already observed in several papers, varying the number of nodes degrades the performance (columns~(a) and~(d)). Interestingly, varying the number of modes only also has a negative impact (columns~(b) and~(e)), and the same holds when varying the scaling of the node coordinates in the TSP (column~(c)) or the vehicle capacity in the CVRP (column~(f)). Similar results of performance degradation on generalization of the GCN model are given in Table~\ref{tab:gen:GCN-TSP} for TSP. These results confirm the drastic lack of generalization between the models, even on seemingly closely related instance distributions. In the next section, we propose an approach to tackle this problem.

\begin{table*}[t]
\centering
\small
\begin{tabular}{l|rrr|l|rrr|l|rrr}
     $\scriptsize{N\frac{\textrm{test}\rightarrow}{\textrm{train}\downarrow}}$ & $N{=}20$ & $N{=}50$ & $N{=}100$ & $\scriptsize{M\frac{\textrm{test}\rightarrow}{\textrm{train}\downarrow}}$ & $M{=}0$ & $M{=}3$ & $M{=}6$ & $\scriptsize{L\frac{\textrm{test}\rightarrow}{\textrm{train}\downarrow}}$ & $L{=}1$ & $L{=}5$ & $L{=}10$\\
    \hline
    $N{=}20$ & \textbf{0.08} & 1.78 & 22.61 & $M{=}0$ & \textbf{1.47} & 32.17 & 2.74 & $L{=}1$ & \textbf{1.48} & 282.55 & 292.39 \\
    $N{=}50$ & 0.35 & \textbf{0.52} & 2.95 & $M{=}3$ & 26.38 & \textbf{1.86} & 7.32 & $L{=}5$ & 32.84 & \textbf{1.44} & 13.83\\
    $N{=}100$ & 3.78 & 2.33 & \textbf{2.26} & $M{=}6$ & 6.91 & 6.01 & \textbf{2.0} & $L{=}10$ & 98.62 & 7.12 & \textbf{1.53} \\
        \hline
    \multicolumn{4}{c|}{(a) $N$ ($M{=}0,L{=}1$)} &
     \multicolumn{4}{c|}{(b) $M$ ($N{=}40,L{=}1$)} &
     \multicolumn{4}{c}{(c) $L$ ($N{=}40,M{=}0$)}\\
\end{tabular}
\\
\begin{tabular}{l|rrr|l|rrr|l|rrr}
    $\scriptsize{N\frac{\textrm{test}\rightarrow}{\textrm{train}\downarrow}}$ & $N{=}20$ & $N{=}50$ & $N{=}100$ &
    $\scriptsize{M\frac{\textrm{test}\rightarrow}{\textrm{train}\downarrow}}$ &
    $M{=}1$ & $M{=}3$ & $M{=8}$ &
    $\scriptsize{C\frac{\textrm{test}\rightarrow}{\textrm{train}\downarrow}}$ & $C{=}20$ & $C{=}30$ & $C{=}50$\\

    \hline
    $N{=}20$ & \textbf{4.52} & 12.61 & 20.23 & $M{=}1$ & \textbf{4.39} & 51.02 & 102.07 & $C{=}20$ & \textbf{5.83} & 8.25 & 12.23 \\

    $N{=}50$ & 7.99 & \textbf{6.93} & 8.47 & $M{=}3$ & 5.67 & \textbf{6.32} & 16.14 & $C{=}30$ & 6.13 & \textbf{7.37} & 9.39\\

    $N{=}100$ & 12.90 & 9.75 & \textbf{7.11} & $M{=}8$ & 14.91 & 8.67 & \textbf{7.85} & $C{=}50$ & 12.27 & 8.56 & \textbf{7.99} \\

    \hline
    \multicolumn{4}{c|}{(d) $N$ ($M{=}0,C{=}\textrm{func}(N)$)} &
    \multicolumn{4}{c|}{(e) $M$ ($N{=}50,C{=}40$)} &
     \multicolumn{4}{c}{(f) $C$ ($N{=}\textrm{func}(C),M{=}0$)}\\
\end{tabular}
 \caption{\label{gen:AM_optgap}\textbf{Performance deterioration of AM(TSP and CVRP)}: Average gap of the AM model (in percentage, over 5000 test instances) when trained and tested on TSP instances with different (a) number of nodes $N$ (b) number of modes $M$ and (c) scale $L$; and CVRP instances with different (d) number of nodes $N$, (e) number of modes $M$  and (f) vehicle capacities $C$.}
\begin{tabular}{l|rrr|l|rrr|l|rrr}
     $N\frac{\textrm{test}\rightarrow}{\textrm{train}\downarrow}$ & $N{=}20$ & $N{=}50$ & $N{=}100$ & $M\frac{\textrm{test}\rightarrow}{\textrm{train}\downarrow}$ & $M{=}0$ & $M{=}3$ & $M{=}8$ & $L\frac{\textrm{test}\rightarrow}{\textrm{train}\downarrow}$ & $L{=}1$ & $L{=}5$ & $L{=}10$\\
    \hline
    $N{=}20$ & \textbf{1.83} & 38.66 & 77.31 & $M{=}0$ & \textbf{5.05} & 35.86 & 26.01 & $L{=}1$ & \textbf{5.10} & 28.15 & 32.46 \\
    $N{=}50$ & 22.05 & \textbf{5.10} & 43.76 & $M{=}3$ & 35.40 & \textbf{6.96} & 28.71 & $L{=}5$ & 272.58 & \textbf{5.23} & 25.41\\
    $N{=}100$ & 43.86 & 37.26 & \textbf{14.79} & $M{=}8$ & 32.74 & 36.29 & \textbf{5.48} & $L{=}10$ & 289.51 & 66.28 & \textbf{5.46} \\
        \hline
    \multicolumn{4}{c|}{(a) $N$ ($M{=}0,L{=}1$)} &
     \multicolumn{4}{c|}{(b) $M$ ($N{=}50,L{=}1$)} &
     \multicolumn{4}{c}{(c) $L$ ($N{=}50,M{=}0$)}\\
\end{tabular}
\caption{\label{tab:gen:GCN-TSP} \textbf{Performance deterioration of GCN(TSP)}: Average gap of the GCN model, when varying (a) the number of nodes $N$ (b) the number of modes $M$ and (c) the scale $L$.}
\end{table*}
\section{Meta-learning of NCO heuristics}
\label{sec:meta}
\begin{figure}[ht]
\centering
\includegraphics[width=1\columnwidth]{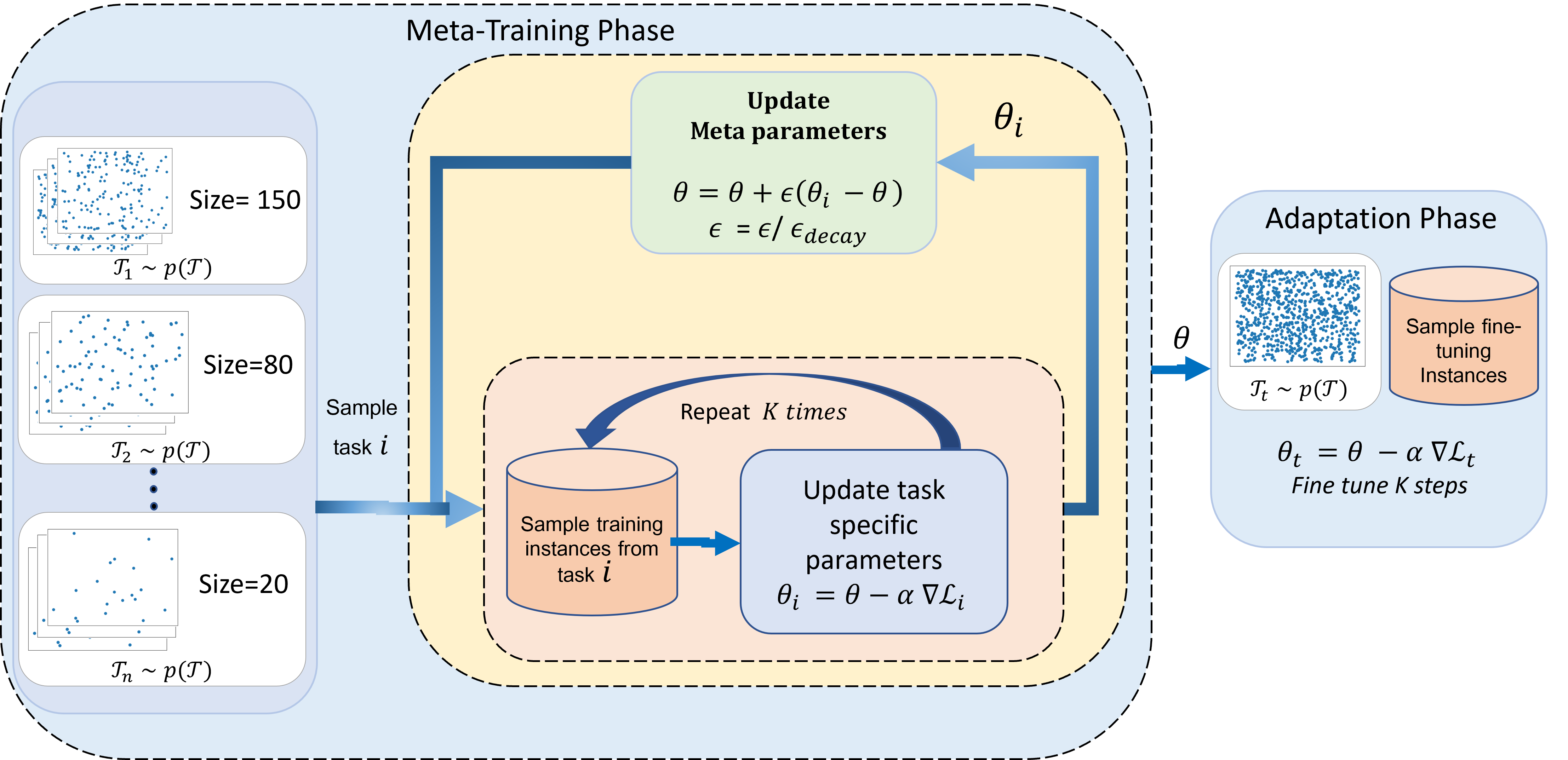}
\caption{\label{fig:Arch}Architectural diagram of our proposed method. Note that in the training phase of the figure, instead of size variation, one can have different types of distribution shifts.}
\end{figure}
The goal of this paper is to introduce an NCO approach capable of out-of-distribution generalization for a given CO problem. Since NCO methods tend to perform well on fixed instance distributions, our strategy to promote out-of-distribution generalization is to modify the way the model is trained without changing its architecture.

Concretely, given a CO problem (e.g. the TSP), we assume that we have a prior over the relevant tasks (instance distributions), possibly based on historical data. For instance, we may know that the customers in our TSP are generally clustered around city centers, but without knowing how many clusters. Our underlying assumption is that it easier and more realistic to obtain a prior distribution on target tasks, rather than the target task itself.
We propose to first train a model to learn an efficient heuristic on a sample of tasks (e.g. TSP instances with different numbers of modes). Then, considering a new unseen task (unseen number of modes), we would use a \textit{limited number of samples} (few-shots) from that task to specialize the learned heuristic and maximize its performance on it.
Fig.~\ref{fig:Arch} illustrates our proposed approach.

Formally, given an NCO model with parameters $\theta$ and a distribution of tasks $\mathcal{T}$, our goal is to compute a parameter $\theta$ such that, given an unseen task $t \sim \mathcal{T}$ with associated loss $\mathcal{L}_t$, after $K$ gradient updates, the fine-tuned parameter minimizes $\mathcal{L}_t$, i.e.
\begin{equation}
    \label{eq:meta-pb}
    \min_\theta \,
    \mathbb{E}_{t \sim \mathcal{T}} [\mathcal{L}_t (\theta_t^{(K)})],
\end{equation}
where $\theta_t^{(K)}$ is the fine-tuned parameter after $K$ gradient updates of $\theta$ using batches of instances from task $t$.
Problem~\eqref{eq:meta-pb} can be viewed as a few-shot meta-learning optimization problem.
We approach it in a model-agnostic fashion by leveraging the generic \textsc{Reptile} meta-learning algorithm~\cite{nichol_first-order_2018}. Given a task distribution, \textsc{Reptile} is a surprisingly simple algorithm to learn a model that performs well on unseen tasks of that distribution. Compared to the seminal MAML framework~\cite{finn_model-agnostic_2017}, \textsc{Reptile} is a first-order method that does not differentiate through the fine-tuning process at train time, making it feasible to work with higher values of $K$. And we observed experimentally that in our context, to fine-tune a model to a new task, we need up to $K=50$ steps, which is beyond MAML's practical limits. Furthermore, since \textsc{Reptile} uses only first-order gradients with a very simple form, it is more efficient, both in terms of computation and memory.
Using \textsc{Reptile}, we meta-train each model on the given task distribution to obtain an effective initialization of the parameters, which can subsequently be adapted to a new target task using a limited number of fine-tuning samples from that task.

The first step to optimize Eq.~\ref{eq:meta-pb} consists of $K$ updates of task specific parameters for a task $\mathcal{T}_i \sim \mathcal{T}$ as follows:
\begin{equation}
\begin{gathered}
 \theta_i^0 = \theta, \\
 \theta_{i}^j=\theta_{i}^{j-1} -\alpha \nabla_{\theta_i^{j-1}}\mathcal{L}_i, \;\;\; \forall j \in [1 \ldots K].
\end{gathered}
\end{equation}
In the above equation, the hyper-parameter $\alpha$ controls the learning rate. Then, using the updated parameters $\theta_{i}^K$ obtained at the end of the $K$ steps, we update the meta-parameter $\theta$ as follows:
\begin{equation}
    \label{eq:meta_gradient}
    \theta = \theta + \epsilon\left( \theta^K_i - \theta \right).
\end{equation}
This is essentially a weighted combination of the updated task parameters $\theta^K_i$ and previous model parameters $\theta$. The parameter $\varepsilon$ can be interpreted as a step-size in the direction of the Reptile ``gradient'' $\theta^K_i{-}\theta$. It controls the contribution of task specific parameters to the overall model parameters. We iterate over $\mathcal{T}_i \sim \mathcal{T}$ by computing Eq.~\ref{eq:meta_gradient} for different tasks and then using it for optimizing Eq.~\ref{eq:meta-pb}.

\noindent \\
\textbf{Scheduling $\varepsilon$: first specialize then generalize}. As mentioned above, parameter $\varepsilon$ controls the contribution of the task specific loss to the global meta parameters $\theta$ update in Eq.~\ref{eq:meta_gradient}. A high value of $\varepsilon$ leads to overfitting on the training task while a low value leads to inefficient learning of the task itself. In order to tackle such scenario, in this work we utilize a simple decaying schedule for $\varepsilon$ which starts close to 1 (i.e. the meta-parameters are updated to the fine-tuned ones) and tends to 0 as the training proceeds, thus stabilizing the meta-parameter that is more likely to work well for all tasks.

\noindent \\
\textbf{Fine-tuning for target adaptation:} Once the model is meta-trained on a diverse set of tasks, given a new task $\mathcal{T}_t$, we initialize the target task parameters to the value of the meta-trained model $\theta$ and do a number a fine-tuning steps to get the specialized model for the new task. Essentially,
\begin{equation}
\begin{gathered}
 \theta_t^0 = \theta, \\
 \theta_{t}^j=\theta_{t}^{j-1} -\alpha \nabla_{\theta_t^{j-1}}\mathcal{L}_t, \;\;\; \forall j \in [1 \ldots K].
\end{gathered}
\end{equation}

We can now detail the meta-training procedure of NCO models for the TSP and CVRP problems over our two state-of-the-art reinforcement learning (AM) and supervised learning (GCN) approaches to NCO heuristic learning. For simplicity, we use as default problem the TSP in this section, while the adaptation of the algorithms for the CVRP is presented in Sec.~\ref{sec:meta-cvrp-sup} in Supplementary Material.

\subsection{Meta-learning of RL-based NCO heuristics (AM model)}

The RL based model of~\cite{kool_attention_2019} (AM) consists of learning a policy that takes as input a graph representing the TSP instance and outputs the solution as a sequence of graph nodes. The policy is parameterized by a neural network with attention based encoder and decoder~\cite{vaswani_attention_2017} stages. The encoder computes nodes and graph embeddings; using these embeddings and a context vector, the decoder produces the sequence of input nodes in an auto-regressive manner.
In effect, given a graph instance $G$ with $N$ nodes, the model produces a probability distribution $\pi_\theta(\sigma | G)$ from which one can sample to get a full solution in the form of a permutation $\sigma = (\sigma_1, \dots, \sigma_N)$ of $\{1, \dots, N\}$. The policy parameter $\theta$ is optimized to minimize the loss: {\footnotesize $\mathcal{L}(\theta | G) = \mathbb{E}_{\pi_\theta(\sigma | G)}[c(\sigma)]$},
where $c$ is the cost (or length) of the tour $\sigma$. The REINFORCE~\cite{williams_simple_1992} gradient estimator is used: {\footnotesize $\nabla_\theta \mathcal{L}(\theta | G) = \mathbb{E}_{\pi_\theta(\sigma | G)} [(c(\sigma) - b(G)) \nabla_\theta \log \pi_\theta(\sigma|G)]$}.
As in~\cite{kool_attention_2019}, we use as baseline $b$ the cost of a greedy rollout of the best model policy, that is updated periodically during training.
\noindent \\
\textbf{Meta-training of AM:}
Algorithm~\ref{alg:meta-am} describes our approach for meta-training the AM model for the TSP problem. For simplicity, the distribution of tasks that we consider here is uniform over a finite fixed set of tasks. Otherwise, one just needs to define the task-specific baseline parameters $\theta_i^\text{BL}$ on the fly when a task is sampled for the first time. The training consists of repeatedly sampling a task (line~3), doing $K$ update of the meta-parameters $\theta$ using samples from that task to get fine-tuned parameters $\theta_i$, then updating the meta-parameters as a convex combination of their previous value and the fine-tuned value (line~15). Note that the baseline need not be updated at each step (line~12), but only periodically, to improve the stability of the gradients.
\begin{algorithm}[tb]
\caption {Meta-training of the Attention Model}
\label{alg:meta-am}
\begin{algorithmic} [1]
    \REQUIRE Task set $\mathcal{T}$, \# updates $K$, threshold $\beta$, step-size initialization $\varepsilon_0 \approx 1$ and decay $\varepsilon_{\text decay} > 1$
	\STATE Initialize meta-parameters $\theta$ randomly, baseline parameters $\theta^b_i = \theta$ for $\mathcal{T}_i \in \mathcal{T}$ and step-size $\varepsilon = \varepsilon_0$
	\WHILE{not done}
	    \STATE Sample a task $\mathcal{T}_i \in \mathcal{T}$
	    \STATE Initialize adapted parameters $\theta_i \leftarrow \theta$
 	    \FOR{$K$ times}
    	    \STATE Sample batch of graphs $g_k$ from task $\mathcal{T}_i$
    	    \STATE $\sigma_k \leftarrow \textrm{SampleRollout}(g_k, \pi_{\theta_i}) \quad \forall k$
    	    \STATE $\sigma_k^b \leftarrow \textrm{GreedyRollout}(g_k, \pi_{\theta_i^b}) \quad \forall k$
    	    \STATE $\nabla_\theta \mathcal{L}_i \leftarrow \sum_k (c(\sigma_k) - c(\sigma_k^b)) \nabla_\theta \log \pi_{\theta_i}(\sigma_k)$
    	    \STATE $\theta_i \leftarrow \textrm{Adam}(\theta_i, \nabla_\theta \mathcal{L}_i)$ \COMMENT{Update for task $\mathcal{T}_i$}
    	\ENDFOR
    	\IF{OneSidedPairedTTest($\pi_{\theta_i}, \pi_{\theta_i^b}$) $< \beta$}
    	    \STATE Update baseline $\theta_i^b \leftarrow \theta_i$ \COMMENT{Update task specific baseline}
    	\ENDIF
    \STATE Update $\theta \leftarrow (1 - \varepsilon)\theta + \varepsilon \theta_i$, $\;\varepsilon \leftarrow \varepsilon/\varepsilon_{\text decay}$   \COMMENT{Update meta parameters, step size}
    \ENDWHILE
\end{algorithmic}
\end{algorithm}

\subsection{Meta-learning of supervised NCO heuristics (GCN model)}

The supervised model of~\cite{joshi_efficient_2019} (GCN) consists of a Graph Convolution Network that takes as input a TSP instance as a graph $G$ and outputs, for each edge $ij$ in $G$, predicted probabilities $\hat{s}_{ij}$ of being part of the optimal solution. It is trained using a weighted binary cross-entropy loss between the predictions and the ground-truth solution $s_{ij}$ provided by the exact solver Concorde~\cite{applegate_traveling_2011}:
\hspace{-0.1in}\begin{equation}
 \label{eq:gcn-loss}
     \mathcal{L}(\theta|G) = \sum_{ij \in G} w_0 s_{ij} \log(\hat{s}_{ij}) + w_1 (1 - s_{ij}) \log(1 - \hat{s}_{ij}),
\end{equation}
where $w_0$ and $w_1$ are class weights meant to compensate the inherent class imbalance, and $B$ is the batch size. The predicted probabilities are then used either to greedily construct a tour, or as an input to a beam search procedure. For simplicity, and because we are interested in the learning component of the method, we only consider here the greedy version.
\noindent \\
\textbf{Meta-training of GCN:} Algorithm~\ref{alg:meta-gcn} in Supplementary Material describes our approach for meta-training the GCN model. In contrast to Algorithm~\ref{alg:meta-am}, we need here to fix the training tasks since the ground-truth optimal solutions must be precomputed in this supervised learning framework.

\section{Experiments}
\label{sec:exp}
The goal of our experiments is to demonstrate the effectiveness of meta-learning for achieving generalization in NCO.
More precisely, given a prior distribution of tasks, we aim to answer the following questions:
(i) How does the (fine-tuned) meta-trained NCO models perform on unseen tasks, in terms of \textit{optimality gaps} and \textit{sample efficiency}?
(ii) How does the meta-trained models perform on unseen tasks that are \textit{interpolated} or \textit{extrapolated} from the training tasks?
(iii) How effective is our proposed \textit{decaying step-size} strategy in the Reptile meta-learning algorithm for our NCO tasks?

\noindent \\
\textbf{Experimental setup.}
Experiments were performed on a pool of machines running Intel(R) CPUs with 16 cores, 256GB RAM under CentOS Linux 7, having Nvidia Volta V100 GPUs with 32GB GPU memory. All the models were trained for 24 hours on 1 GPU. The detailed hyperparameters are presented in Sec.~\ref{sec:hyperparam} of the Supp. Mat.. Our code and datasets are available at: \url{https://anonymous.4open.science/r/meta-NCO}

\noindent \\
\textbf{Task distributions.}
For the TSP (resp. CVRP) experiments, we consider four task distributions (Section~\ref{sec:graph-distrib}) which are obtained from  $\task{N=40,M=0,L=1}$ (resp. $\task{N=50,M=0,C=40,L=1}$) as follows:
(i) a {\textit var-size} distribution is obtained by varying {\scriptsize $N$} only, and for training tasks within this distribution we use {\scriptsize $N{\in}\{10,20,30,50\}$}; (ii) {\textit var-mode} distribution by varying {\scriptsize $M$} only, and for training {\scriptsize $M{\in}\{1,2,5\}$}; (iii) {\textit mixed-var} distribution by varying both {\scriptsize $N$} and {\scriptsize $M$} and training with {\scriptsize $(N,M) \in \{20,30,50\} \times \{1,2,4\}$}; and (iv) only for CVRP: {\textit var-capacity} distribution by varying {\scriptsize $C$} only, for training {\scriptsize $C{\in}\{10,30,40\}$}. As test tasks, we use values that are both within the training tasks range to evaluate the {\textit interpolation} performance (e.g. {\scriptsize $M{=}3$} for (ii)) and outside to evaluate the  {\textit extrapolation} performance (e.g. {\scriptsize $N{=}100$} for (i)). More details about the distributions are presented in Sec.~\ref{sec:task-dist-sup} of the Supp. Mat.

\noindent \\
\textbf{Datasets.}
We generate synthetic TSP and CVRP instances, according to the previously described task distributions. For AM training, samples are generated on demand while for the GCN model, we generate for each task a training set of 1M instances, a validation and test set of 5K instances each and use the Concorde solver~\cite{applegate_traveling_2011} and LKH \cite{helsgaun_extension_2017} to get the associated ground-truth solutions for TSP and CVRP respectively (as was done in the original work). In order to fine-tune the meta-trained models, we sample a set of instances from the new task, containing either 3K (AM) or 1K (GCN) samples; these numbers were chosen as approximately $0.01\%$ and $0.1\%$ of the number of samples used during the 24 hours training of the AM and GCN models respectively (see details in Sec.~\ref{sec:sample-eff-sup} in Supp. Mat.).
In addition to synthetic datasets, we evaluate our models on the realistic datasets: TSPlib and CVRPlib. The precise settings and results are presented in Section~\ref{sec:realistic-datasets}.

\noindent \\
\textbf{Models.}
We use the AM-based heuristics of~\cite{kool_attention_2019} for TSP and CVRP. For the GCN model, we use the model provided by~\cite{joshi_efficient_2019} for the TSP and its adaptation by~\cite{kool_deep_2021} for the CVRP.
For a given task distribution (e.g. {\textit variable-size}) we consider the following models:
\begin{itemize}
    \item {\tt meta-AM} (resp. {\tt meta-GCN}): the AM (resp. GCN) model meta-trained (following Algorithm~\ref{alg:meta-am} or~\ref{alg:meta-gcn} for TSP). E.g. for the {variable-size} distribution, we denote this model {\tt meta-AM-N} (resp. {\tt meta-GCN-N}).
    \item {\tt multi-AM} (resp. {\tt multi-GCN}): the AM (resp. GCN) model trained with instances coming equiprobably from the training tasks. E.g. for the {variable-mode} distribution, we denote this model {\tt multi-AM-M} (resp. {\tt multi-GCN-M}).
    \item {\tt oracle-AM} (resp. {\tt oracle-GCN}): original AM (resp. GCN) model trained on the {\textit test} instance distribution, that is unseen during training of both the meta and multi models. Note that the meta-models are not meant to improve over the oracles' performance, although we will see that it happens sometimes.
\end{itemize}
To simplify the notations, we only explicitly differentiate between TSP and CVRP if it is not clear from the context. Since we are interested in the generalization of the neural models, regardless of the final decoding step (greedy, sampling, beam-search, etc), we use a simple greedy decoding for all the models. Besides, because our training is restricted to 24 hours for all the models (which is sufficient to ensure convergence of the training, see Fig~\ref{fig:AM-tsp-time} of Supp. Mat.), the results may not be as good are those reported in the original papers. To evaluate the impact of the meta-training on generalization when everything else fixed, we focus on the relative gap in performance between the different models.

\noindent \\
\textbf{Generalization performance:}
To evaluate the generalization ability of the meta-trained models, we present in Table~\ref{tab:gen-big-table} the performance of the different models at 0-shot generalization ($K$=0) and after $K$=50 fine-tuning steps, for various pairs of prior task distributions and unseen test tasks. We observe that
in all cases the fine-tuned {\tt meta-AM} clearly outperforms the fine-tuned baseline {\tt multi-AM} and even outperforms the {\tt oracle-AM} model in 7 out of 12 tasks.
Similar observations hold for the {\tt meta-GCN} model: it is better both at 0-shot generalization and after fine-tuning than the {\tt multi-GCN} baseline, and it outperforms the oracle in 2 out of 6 tasks. These results show that {\tt meta-AM} is able to achieve impressive quality while using a negligible amount of training data of the target task compared to the original model ({\tt oracle-AM}). More results on different target tasks as well as plots of the evolution of the performance with the number of fine-tuning steps are presented in Sec.~\ref{sec:additional-results} of the Supp. Mat.
\begin{table*}[t!]
\centering

\begin{tabular}{|l|l||rr|rr|rr|}
\hline
\multirow{7}{*}{\rotatebox[origin=c]{90}{\texttt{TSP}}} & \multicolumn{1}{c|| }{Tasks $\rightarrow$}  & \multicolumn{2}{c|}{{\textit var-size} distrib.} & \multicolumn{2}{c|}{{\textit var-mode} distrib.} &
\multicolumn{2}{c|}{{\textit mixed-var} distrib.} \\
& \multicolumn{1}{c||}{Models $\downarrow$} & \multicolumn{1}{c}{{\scriptsize N}=100} & \multicolumn{1}{c|}{{\scriptsize N}=150} & \multicolumn{1}{c}{{\scriptsize M}=3} & \multicolumn{1}{c|}{{\scriptsize M}=8} & \multicolumn{1}{c}{{\scriptsize (N,M)}=(40,6)} & \multicolumn{1}{c|}{{\scriptsize (N,M)}=(40,8)} \\
\cline{2-8}
& {\tt oracle-AM} & 5.96\% & 12.08\% & 1.87 \% & 1.83\% & 2.00\% & 1.83\%  \\
\cline{2-8}
& Farthest Ins.\cite{rosenkrantz_analysis_2009} & 7.48\% & \textbf{8.55}\% & 2.08\% & 2.27\% & 16.32\% & 11.70\%  \\
& {\tt multi-AM} ({\scriptsize $K$}=0) & 8.73\% & 14.40\% & 5.57\% & 6.20\% & 10.70\% & 15.18\%  \\
& {\tt multi-AM} ({\scriptsize $K$}=50) & 7.25\% & 10.87\% & 5.26\% & 4.60\% & 7.59\% & 10.26\%  \\
& {\tt meta-AM} ({\scriptsize $K$}=0) & 7.10\% & 12.25\% & 1.96\% & 2.16\% & 2.41\% & 3.50\% \\
& {\tt meta-AM} ({\scriptsize $K$}=50) &\textbf{5.58\%} & 9.84\% & \textbf{1.82\%} & \textbf{1.70\%} & \textbf{2.15\%} & \textbf{2.93\%}  \\
\hline 
\multirow{6}{*}{\rotatebox[origin=c]{90}{\texttt{CVRP}}} & \multicolumn{1}{c|| }{Tasks $\rightarrow$}  & \multicolumn{2}{c|}{{\textit var-size} distrib.} & \multicolumn{2}{c|}{{\textit var-mode} distrib.} &
\multicolumn{2}{c|}{{\textit var-capacity} distrib.} \\
& \multicolumn{1}{c||}{Models $\downarrow$} & \multicolumn{1}{c}{{\scriptsize N}=100} & \multicolumn{1}{c|}{{\scriptsize N}=150} & \multicolumn{1}{c}{{\scriptsize M}=3} & \multicolumn{1}{c|}{{\scriptsize M}=8} & \multicolumn{1}{c}{{\scriptsize C}=20} & \multicolumn{1}{c|}{{\scriptsize C}=50} \\
\cline{2-8}
& {\tt oracle-AM} & 8.71\% & 11.56\% & 6.32 \% & 7.85\% & 5.83\% & 8.01\% \\ \cline{2-8}
& {\tt multi-AM} ({\scriptsize $K$}=0) & 18.82\% & 18.76\% & 7.87\% & 12.65\% & 9.15\% & 14.28\%  \\
& {\tt multi-AM} ({\scriptsize $K$}=50)  & 9.18\% & 11.41\% & 7.58\% & 10.20\% & 8.09\% & 10.16\%  \\
& {\tt meta-AM }({\scriptsize $K$}=0) & 11.50\% & 16.42\% & 6.05\% & 9.38\% & 6.26\% & 8.94\% \\
& {\tt meta-AM} ({\scriptsize $K$}=50)  &\textbf{7.71\%} &\textbf{ 9.91\%} & \textbf{5.96\% }& \textbf{8.45\% }& \textbf{6.05\%} &  \textbf{8.82\%} \\

\hline \arrayrulecolor{white}\hline \hline  \hline  \arrayrulecolor{black}\hline

\multirow{6}{*}{\rotatebox[origin=c]{90}{\texttt{TSP}}} & \multicolumn{1}{c|| }{Tasks $\rightarrow$}  & \multicolumn{2}{c|}{{\textit var-size} distrib.} & \multicolumn{2}{c|}{{\textit var-mode} distrib.} &
\multicolumn{2}{c|}{{\textit mixed-var} distrib.} \\
& \multicolumn{1}{c||}{Models $\downarrow$} &  \multicolumn{1}{c}{{\scriptsize N}=80} & \multicolumn{1}{c|}{{\scriptsize N}=100} & \multicolumn{1}{c}{{\scriptsize M}=3} & \multicolumn{1}{c|}{{\scriptsize M}=8} & \multicolumn{1}{c}{{\scriptsize (N,M)}=(40,6)} & \multicolumn{1}{c|}{{\scriptsize (N,M)}=(40,8)} \\
\cline{2-8}
& {\tt oracle-GCN} & 12.34\% & 14.72\% & 7.65\% & 6.21\% & 6.06\% & 3.22\%  \\
\cline{2-8}
& {\tt multi-GCN} ({\scriptsize K}=0) & 28.40\% & 34.29\% & 9.22\% & 7.89\% & 28.01\% & 5.05\%  \\
& {\tt multi-GCN} ({\scriptsize K}=50) & 16.73\% & 30.80\% & 8.43\% & 6.59\% & 5.99\% & 4.42\%  \\
& {\tt meta-GCN} ({\scriptsize K}=0) & 19.70\% & 32.01\% & 8.19\% & 7.32\% & 6.62\% & 3.72\% \\
& {\tt meta-GCN} ({\scriptsize K}=50) & \textbf{13.73\%} &\textbf{ 18.42\%} & \textbf{7.72\%} &\textbf{ 6.45\%} & \textbf{5.67\%} & \textbf{3.17\%}  \\
\hline

\end{tabular}
\caption{Average optimality gaps over 5,000 instances of the target tasks (e.g. N=100) coming from different prior task distributions (e.g. {\textit var-size} distribution). {\tt oracle-AM/GCN} denote the AM/GCN models trained on the target task. {\tt multi-AM/GCN} and {\tt meta-AM/GCN} are trained on a set of tasks from the prior distribution that does not contain the target tasks. $K$ is the number of fine-tuning steps. In bold: for each model (AM or GCN) and each problem (TSP or CVRP), the best generalization result among the methods that were not trained on the target task.}
\label{tab:gen-big-table}
\end{table*}

\noindent \\
\textbf{Time and sample efficiency.} \label{sec:sample-time-main}For a complete evaluation of the proposed meta-training and then fine-tuning approach for NCO, we discuss here its cost in terms of the  fine-tuning time and number of training samples from the target task required to reach the optimality gaps of Table~\ref{tab:gen-big-table}. Regarding the fine-tuning time, the 50 fine-tuning steps took 2 to 6m for {\tt meta-AM} and 43s to 2m for {\tt meta-GCN}. Further, generating the 1k optimal solutions for fine-tuning the supervised {\tt meta-GCN model} took up to 17m for TSP150 and 20h for CVRP150. These values should be compared to the generation time of the 1M solutions for training the {\tt oracle-GCN} model on the target instance distribution. Besides, for example for TSP with {\scriptsize $M{=}3$}, we observed that {\tt oracle-AM} needs around 23 hours and more than $30$ Million samples of the target task to reach the optimality gap of $1.82\%$. On the other hand, {\tt meta-AM-M} only used $3000$ samples from the target task and achieved a better performance after a few fine-tuning steps and less than 6 minutes. The baseline approach {\tt multi-AM-M} was still far away at $5.2\%$ optimality gap after fine-tuning. Similar observations hold for {\tt meta-GCN} on TSP with {\scriptsize $M{=}3$}: {\tt{Oracle-GCN-M}} needs around 22 hours and 1 Million instances of labeled data (with optimal solutions) to reach an optimality gap of 7.72\%, while {\tt meta-GCN-M} reaches the same performance in just 16 seconds, using 500 solved instances. Hence, one model trained using our prescribed meta-learning approach can be used to adapt to different tasks efficiently within a short span of time and using few fine-tuning samples. More details on training time and number of samples used for different tasks can be found in the Table~\ref{tab:datasamples} of the Supp. Mat. Additionally, Fig.~\ref{fig:AM-tsp-time} in the  Supp. Mat. presents the performance of different models w.r.t time on test tasks during their course of training.

\subsection{Experiments on real-world datasets}
\label{sec:realistic-datasets}
To evaluate the performance of our approach beyond synthetic datasets, we ran experiments on two well-established OR datasets: {TSPlib}\footnote{http://comopt.ifi.uni-heidelberg.de/software/TSPLIB95/} and {CVRPlib}\footnote{http://vrp.atd-lab.inf.puc-rio.br/index.php/en/}. From  TSPlib we took the $28$ instances of size $50$ to $200$ nodes. Note that in this context, the RL approach which does not rely on labeled data for fine-tuning is more appropriate. Since these instances are heterogeneous (i.e. no clear underlying distribution), we directly fine-tune the models on each test instance.
This is an extreme case of our setting where the target task is reduced to 1 instance. We tested the models that were (meta-)trained on the {\textit variable-size} distribution of synthetic instances for {\tt meta-AM} and {\tt multi-AM}. For AM we took the pretrained model on graphs of size $N$=100.
Because of space limitation, we grouped the instances per size range and report in Table~\ref{tab:tsp-vrp-lib-results} the average optimality gap obtained after $K$=100 fine-tuning steps, taking $20$s to $1$m (detailed per-instance results in Sec.~\ref{TSPLIB:expts} of Supp Mat). Note that in this case we also fine-tune the AM model since it was not trained on the target instances distribution.

From CVRPlib we used the 106 instances of size up to 200 nodes. Since instances are grouped by sets, we apply our few-shot learning setting: fine-tuning for $50$ steps on approximately 10\% of the instances of a set and testing on the rest. In Table~\ref{tab:tsp-vrp-lib-results}, we report the average optimality gap over 5 random fine-tuning/test splits for each set. The results are consistent with our previous observations and illustrate the superior performance of our proposed meta-learning strategy in this realistic setting. It also shows that even if the prior task distribution is not perfect (in the sense that it does not include the target task), the meta-training gives a strong parameter initialization which one can fine-tune effectively on the target task.

\begin{table}
\small
\centering
\begin{tabular}{|c|rrr||rrrrr|}
\hline
\multirow{2}{*}{$\frac{\textrm{Dataset}\rightarrow}{\textrm{Model}\downarrow}$} & \multicolumn{3}{c||}{{\textbf TSPlib}} & \multicolumn{5}{c|}{{\textbf CVRPlib}} \\
& $50{-}100$ & $101{-}150$ & $151{-}200$ & Set A & Set B & Set E & Set P & Set X \\
\hline
AM  & $8.52\%$ &  $7.97\% $ & $17.35\%$  & $4.54\%$ & $5.69\%$ &  $31.17\%$&  $5.45\%$ & $12.39\%$  \\
multi-AM  & $11.95\%$ & $13.32\%$ & $26.04\%$ & $5.03\%$ & $5.73\%$ & $\textbf{13.00\%}$ & $6.13\%$ & $15.72\%$  \\
meta-AM & $\textbf{5.95\%}$ & $\textbf{5.91\%}$&  $\textbf{13.22\%}$ & $\textbf{3.56\%}$ & $\textbf{5.07\%}$ & $14.07\%$ & $\textbf{5.03\%}$ &  $\textbf{11.87\%}$  \\
\hline
\end{tabular}
\caption{Average optimality gaps on realistic instances}
\label{tab:tsp-vrp-lib-results}
\end{table}

\subsection{Ablation study}
\textbf{Fixed vs decaying step-size $\varepsilon$.}
In this section, we study the impact of our proposed decaying $\varepsilon$ approach during meta-training. Specifically, Table~\ref{tab:varyalpha} presents the results of using a standard fixed step-size $\varepsilon$ versus a decaying  $\varepsilon$. We see that the decaying $\varepsilon$ version of \texttt{meta-AM} and \texttt{meta-GCN} outperforms the fixed $\varepsilon$ one, both in terms of 0-shot generalization (i.e $K=0$) and after $K{=}50$ steps of fine-tuning. This supports our argument for performing task specialization in the beginning and generalization at the end of the meta-training procedure.

\begin{table*}[h!]
\centering
\begin{tabular}{l|ll|rrrrr|r}
& Test task & Fine-tuning & $\varepsilon{=}0.1$ & $\varepsilon{=}0.3$ & $\varepsilon{=}0.5$ & $\varepsilon{=}0.7$ & $\varepsilon{=}0.9$ & \textit{decaying} $\varepsilon$ \\
\hline
\multirow{4}{*}{\rotatebox[origin=c]{90}{\texttt{meta-AM}}} & \multirow{2}{*}{$N{=}100$}& before ($K{=}0$) & 9.91\% & 8.33\% & 7.52\% & 6.94\% & 6.63\% & \textbf{7.10\%} \\
& & after  ($K{=}50$) & 7.83\% & 6.50\% & 6.03\% & 5.95\% & 5.96\% & \textbf{5.58\%} \\
\cline{2-9}
&\multirow{2}{*}{$M{=}8$} & before ($K{=}0$) & 5.99\% & 3.07\% & 3.38\% & 2.35\% & 2.52\% & \textbf{2.16\%} \\
& & after ($K{=}50$) & 4.78\% & 2.27\% & 2.63\% & 1.87\% & 2.04\% & \textbf{1.70\%}  \\
\hline
\hline
\multirow{4}{*}{\rotatebox[origin=c]{90}{\texttt{meta-GCN}}} & \multirow{2}{*}{$M{=}6$} & before ($K{=}0$) & 13.08\% & 11.90\% & 11.92\% & 12.90\% & 10.11\% & \textbf{6.01\%} \\
& & after  ($K{=}50$) & 9.86\% & 8.27\% & 9.52\% & 10.80\% & 13.16\% & \textbf{5.71\%} \\
\cline{2-9}
&\multirow{2}{*}{$M{=}8$} & before ($K{=}0$) & 9.78\% & 8.81\% & 9.20\% & 11.05\% & 11.96\% & \textbf{7.39\%} \\
& & after  ($K{=}50$) & 8.32\% & 7.37\% & 8.23\% & 9.76\% & 11.80\% & \textbf{6.45\%}  \\
\end{tabular}
\caption{\label{tab:varyalpha}(\textbf{Fixed vs decaying step-size $\varepsilon$)} Average optimality gap, on 5000 TSP instances sampled from a set of test tasks, using the meta-trained models \texttt{meta-AM} (resp. \texttt{meta-GCN}) when trained with a fixed step-size $\varepsilon=\varepsilon_0$ or a ``decaying $\varepsilon$'' where $\varepsilon$ is close to 1 initially and tends to 0 at the end of the training.}
\end{table*}
\section{Conclusion}
In this paper, we address the well-recognized generalization issue of end-to-end NCO methods. In contrast to previous works that aim at having one model perform well on various instance distributions, we propose to learn a model that can efficiently {\textit adapt} to different distributions of instances. To implement this idea, we recast the problem in a meta-learning framework, and introduce a simple yet generic way to meta-train NCO models. We have shown experimentally that our proposed meta-learned RL-based and SL-based NCO heuristics are indeed robust to a variety of distribution shifts for two CO problems. Additionally, the meta-learned models also achieve superior performance on realistic datasets.
We show that our approach can push the boundary of the underlying NCO models by solving instances with up to 200 nodes when the models are trained with only up to 50 nodes. While the known limitations of the underlying models (esp. the attention bottleneck, and fully-connected GCN) prevent tackling much larger problems, our approach could be applied for other models.
Finally note that there are several possible levels of generalization in NCO. In this paper, we have mostly focused on improving the generalization to instance distributions for a fixed CO problem. To go further, one could investigate the generalization to other CO problems. For this more ambitious goal, domain adaptation approaches, which explicitly account for the domain shifts (e.g. using adversarial-based techniques~\cite{tzeng_adversarial_2017}) could be an interesting direction to explore.

\clearpage
\printbibliography

\clearpage
\appendix
\section{Supplementary Material}
\subsection{Meta-training of the GCN model}

Algorithm~2 summarizes the main steps of meta-training the GCN model:
\begin{algorithm}[h]
\caption {Meta-training of the GCN Model}
\label{alg:meta-gcn}
\begin{algorithmic} [1]
    \REQUIRE Task set $\mathcal{T}$, number of updates $K$, step-size initialization $\varepsilon_0 \approx 1$ and decay $\varepsilon_{\text decay} > 1$
	\STATE Initialise meta-parameters $\theta$ randomly, $\varepsilon = \varepsilon_0$
	\WHILE{not done}
	    \STATE Sample a task $\mathcal{T}_i \in \mathcal{T}$
    	\STATE Initialise adapted parameters $\theta_i \leftarrow \theta$
    	\FOR{$K$ times}
    	    \STATE Sample batch of graphs $g_k$ from task $\mathcal{T}_i$
    	    \STATE Compute gradient $\nabla_{\theta}\mathcal{L}_i$ of loss \eqref{eq:gcn-loss} at $\theta_i$,\\
    	     with $\hat{s}_k \leftarrow \textrm{GCN}(g_k, \theta_i) \quad \forall k$

    	    \STATE $\theta_i \leftarrow \textrm{Adam}(\theta_i, \nabla_\theta \mathcal{L}_i)$
    	\ENDFOR
        \STATE Update $\theta \leftarrow (1 - \varepsilon)\theta + \varepsilon \theta_i$, $\;\varepsilon \leftarrow \varepsilon/ \varepsilon_{\text decay}$
    \ENDWHILE
\end{algorithmic}
\end{algorithm}

\subsection{Meta-training for CVRP}
\label{sec:meta-cvrp-sup}
Since the Attention model~[Kool et al., 2019] was proposed for both the TSP and CVRP (and other routing problems) and the GCN model~[Joshi et al., 2019a] originally proposed for the TSP was adapted by~[Kool et al., 2021] to the CVRP, we update the model parameters dimension and input to take into account the depot node, the demand and vehicle capacity, the cost function, greedy rollout baseline and losses as in the original papers. With these updated parameters and functions, Algorithms~1 and~2 remain the same for meta-training the Attention model and the GCN model respectively, for the CVRP.

\subsection{Task distribution datasets}
\label{sec:task-dist-sup}
We summarise the datasets that were generated and that we will make available to hopefully be helpful to the NCO community for evaluating the generalisation ability of their models. In Section~3.1 of the paper, we have described how we generate tasks (or instance distributions) of the form $\task{N,M,C,L}$, for the TSP and the CVRP. With the default values $N{=}50, M{=}0, L{=}1$, we have created three collections of datasets for the TSP:
\begin{itemize}
    \item Varying number of nodes: $N \in \{10, 20, 30, 40, 50, 80, 100, 120, 150\} $
    \item Varying number of modes $M \in \{1, 2, 3, 4, 5, 6, 8 \}$
    \item Varying scale $L \in \{1, 2, 3, 4, 5, 8, 10\}$
\end{itemize}
Similarly, with default values $N{=}50, M{=}0, L{=}1, C{=}40$, we have generated three collections of datasets for the CVRP:
\begin{itemize}
    \item Varying both number of nodes and associated capacity:\\ \hspace{2in} $(N,C) \in \{(10,20), (20,30), (30,35), (50,40), (100,50)\}$
    \item Varying capacity $C \in \{20, 30, 40, 50, 60\}$
    \item Varying number of modes $M \in \{1, 2, 3, 4, 5, 6\}$
\end{itemize}
For each task, we have generated 1M training instances and $2\times10$K validation and test instances. The solutions were computed using the Concorde solver~[Applegate et al., 2006] for TSP and LKH solver~[Helsgaun, 2017] for CVRP. In total, we have generated 16 datasets for the TSP and 16 for the CVRP. Using a 32 cores CPU, it took on average about 10 hours per dataset for TSP and 10 days for the CVRP.

Note that the varying-scale dataset is trivially obtained from the original dataset with scale 1 and the solutions are unchanged. Although one can easily normalise the coordinates in this case and avoid any distribution shift in scale, we have used these datasets as an additional illustration of the sensitivity of the models.

\subsection{Experimental details}

\noindent \\
\textbf{Hyper-parameters:} \label{sec:hyperparam}For both the Attention model and GCN model, we reuse as much as possible the default hyper-parameters of the original papers. For \texttt{meta-AM}, during meta-training we set $K=50$, $\varepsilon_0{=}0.99$ and $\varepsilon_{decay}=1.0003$; \texttt{meta-GCN},during meta-training we set $K=500$, $\varepsilon_0{=}0.99$,  $\varepsilon_{decay}$ between $1.0003$ and $1.25$.
We already discussed the benefit of our proposed  \textit{decaying} $\varepsilon$ parameter in the main paper. Here, we study the impact of the other meta-training specific hyper-parameter $K$ on the performance. Specifically, in Table~\ref{tab:varyK} we show the impact of the number of fine-tuning steps $K$ used during the training of the meta-models \texttt{meta-AM} and \texttt{meta-GCN}, on their test performance.

\begin{table*}[h!]
\centering

\begin{tabular}{l|ll|rrrrrrr|r}
& Test task & Fine-tuning & $K{=}10$ &$K{=}30$ & $K{=}50$ & $K{=}100$ & $K{=}200$ & $K{=}300$ & $K{=}500$  & \textbf{Default}\\
\hline
\multirow{4}{*}{\rotatebox[origin=c]{90}{\texttt{meta-AM}}} &\multirow{2}{*}{$N{=}100$}& before & 10.76\% & 7.41\% & 6.87\% & 6.75\% & \textbf{5.79}\% & 6.05\% & 6.04\% & 7.10\%\\
& & after & 8.13\% & 6.52\% & 5.91\% & 5.92\% & \textbf{5.26}\% & 5.40\% & 5.33\% & 5.58\%\\
\cline{2-11}
&\multirow{2}{*}{$M{=}8$} & before & 3.20\% &3.46\% & 3.16\% & 3.26\% & 2.47\% &2.09\% & 2.99\% &\textbf{2.16\%} \\
& & after & 2.50\% & 2.49\% & 2.43\% & 2.81\% & 2.00\% & 1.76\% & 2.28\%& \textbf{1.70\%}  \\
\hline
\multirow{4}{*}{\rotatebox[origin=c]{90}{\texttt{meta-GCN}}} &\multirow{2}{*}{$M{=}3$}& before & 19.03\% & 16.32\% & 10.32\% & 10.96\% & 10.13\% & 11.70\% & 11.00\% & \textbf{6.01\%} \\
& & after & 19.29\% & 14.44\%  & 9.86\% & 10.53\% & 9.63\% &11.21\% & 10.98\% & \textbf{5.71\%} \\
\cline{2-11}
& \multirow{2}{*}{$M{=}4$} & before & 19.33\% & 13.67\% & 9.28\% & 8.59\% & 10.13\% & 9.85\% & 9.86\% & \textbf{6.42\%} \\
& & after & 17.49\% & 12.69\% & 8.74\% & 8.43\% & 9.71\% & 9.66\% & 9.74\% & \textbf{6.06\%}  \\
\end{tabular}
\caption{\label{tab:varyK} \textbf{Impact of the number of fine-tuning steps \textbf{$K$} used during \textit{meta-training}:} Average optimality gap, on 5000 TSP test instances sampled from a set of test tasks, of the meta-trained models \texttt{meta-AM} and \texttt{meta-GCN}) when \textit{meta-training} is done with a different number of fine-tuning steps $K$ and a fixed step-size $\varepsilon=0.8$. For the \textbf{Default} setting  $\varepsilon_0{=}0.99$, $\varepsilon_{\text decay}{=} 1.0003$ and  during meta-training $K{=}50$ for \texttt{meta-AM} and $K{=}500$ for \texttt{meta-GCN}. Here \textit{before} refers to 0 shot-generalization and \textit{after} refers to value obtained after 50 fine-tuning steps on the  fine-tuning dataset of the Test task.}
\end{table*}

\noindent \\
\subsection{Sample efficiency}
\label{sec:sample-eff-sup}
As mentioned in Sec.~\ref{sec:sample-time-main} of the main paper, the efficiency of our proposed methods {\tt meta-AM} and {\tt meta-GCN} should be viewed keeping in mind the number of training samples of the target task used for training a model from scratch({\tt oracle}). As discussed earlier, for {\tt meta-AM} we use only \textbf{$3000$} samples from the target task during fine-tuning. This is way lower than \textit{millions of samples} used for training a model from scratch({\tt oracle-AM}) as can be seen in Table~\ref{tab:datasamples_TSP}. As we see, in both cases, the number of samples of the target task used by meta-AM are of orders of magnitude smaller than oracle-AM.

\begin{table*}[h!]
\centering
\small
\begin{tabular}{l|p{1cm}|rrrr|rrrr|rr}
      & Scale factor & $N{=}80$ & $N{=}100$ & $N{=}120$ &$N{=}150$ & $M{=}3$ & $M{=}4$ & $M{=}6$ & $M{=}8$  & $L{=}3$ & $L{=}5$ \\
    \hline
    $\#$ oracle-AM & $10^{6}$ & 46.2 & 38.1 & 33.1 & 25.2 & 31.8 & 30.0 & 26.1 & 29.1 & 78.1 & 78.3 \\
     $\frac{\text{\# meta-AM FT}}{\text{\# oracle-AM}}$ & $10^{-5}$ & 6.49 &	7.87 &	9.06 &	11.9 &	9.43 &	10.0 &	11.5 &	10.3 &	3.84 &	3.83
 \\ \hline
        \multicolumn{10}{c}{(a) Travelling Salesman Problem(TSP)  }
    \\
\end{tabular}
\centering
\begin{tabular}{l|p{1cm}|rrrr|rrrr|rr}
     $ $ &Scale factor &$N{=}80$ & $N{=}100$ & $N{=}120$ &$N{=}150$  & $M{=}3$ & $M{=}4$ & $M{=}6$ & $M{=}8$  & $C{=}20$ & $C{=}50$ \\
    \hline
    $\#$ oracle-AM & $10^6$ & 28.2 & 24.6 & 21.0 & 16.0 & 21.2 & 22.1 & 16.0 & 15.0  & 37.3 & 34.1 \\

        $\frac{\text{\# meta-AM FT}}{\text{\#oracle-AM}}$ & $10^{-4}$  & 1.06 &	1.22 &	1.43	 & 1.88 &	1.42 &	1.36 &	1.88 &	2.00 &	0.80 &	0.88
 \\ \hline
        \multicolumn{10}{c}{(b) Capacitated Vehicle Routing Problem(CVRP) }
\end{tabular}
\caption{ \label{tab:datasamples} The first row in each table depicts the number of samples (in millions) of target task used during the training of the original Attention Model {\tt oracle-AM} for different tasks of (a) the TSP and (b) the CVRP. \label{tab:datasamples_TSP} The second row depicts the ratio of number of samples of target task used for training by our proposed model and the oracle model. Note: The \textit{Scale factor} column specifies the multiplier for the number in each cell in that row. For example the value $28.2$ in table (b) at $N{=}80$ is to be read as $28.2 \times 10^{6}$ and the value $1.06$ in table (b) at $N{=}80$ is to be read as $1.06 \times 10^{-4}$.}
\end{table*}

\subsection{Performance vs training time}
\begin{figure}[h]%
    \centering
    \subfloat[\centering Size ]{{\includegraphics[width=5cm]{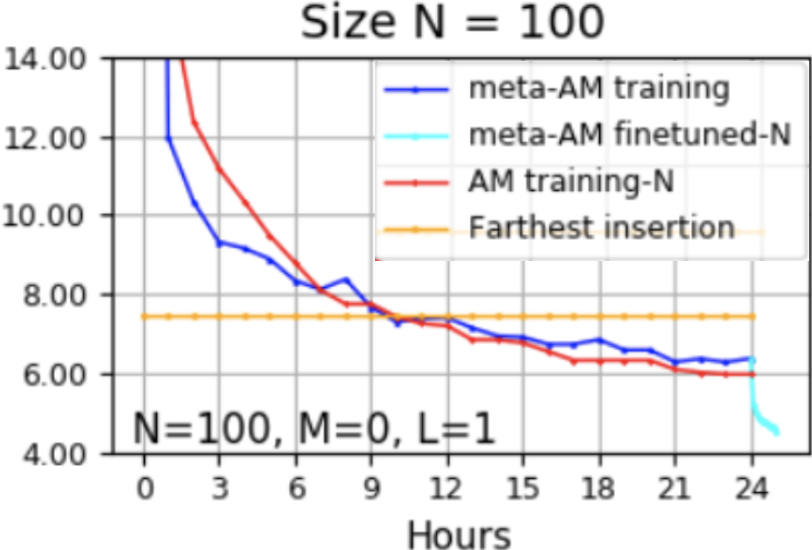} }}%
    \qquad
    \subfloat[\centering Modes ]{{\includegraphics[width=5cm]{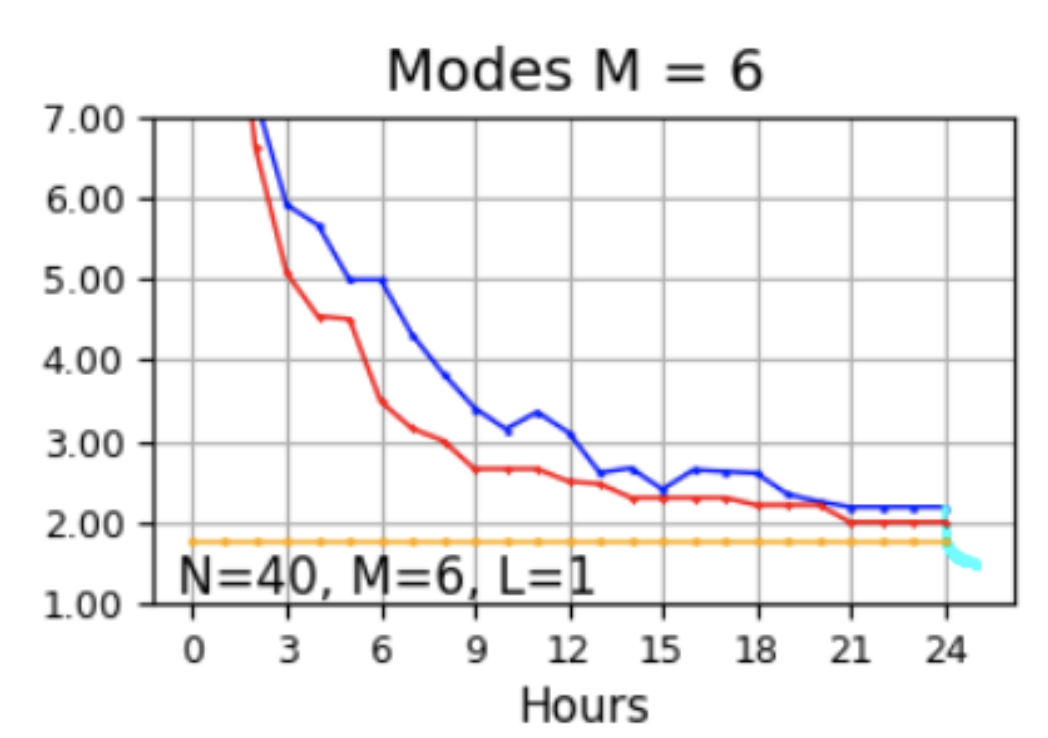} }}%
    \caption{Performance of different models during their training phase, evaluated on test datasets.  }%
    \label{fig:AM-tsp-time}%
\end{figure}

In this section we study the performance of different models evaluated on the test datasets during their course of training. In Fig ~\ref{fig:AM-tsp-time}, we observe that the improvement for all models becomes very slow at the end. Although AM will improve slowly after the 24h mark, the meta fine-tuned meta-model is much faster to reach a better performance.

\noindent \\
\subsection{Additional Results} \label{sec:additional-results}
\begin{figure}[h!]
\centering
\hspace{-0.15in}\includegraphics[scale=0.42]{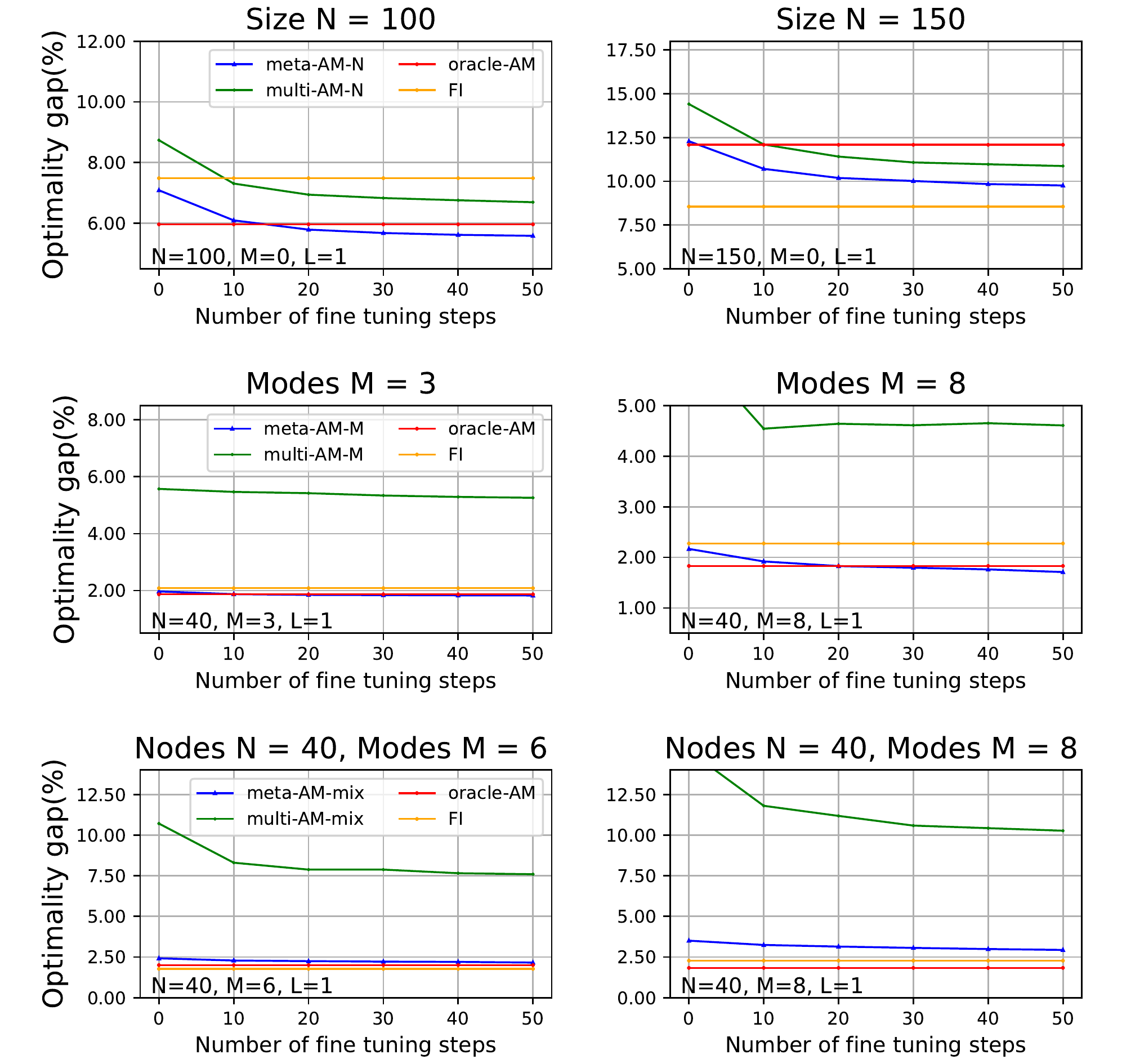}
\caption{\label{fig:meta-vs-multi_tsp} Test performance (average optimality gap, in vertical axis) at different steps of fine-tuning (horizontal axis) of {\tt meta-AM} and {\tt multi-AM}  models on the TSP distribution specified by the parameters at the bottom left corner of each plot. The first row corresponds to {\textit var-size} setting, second  one to {\textit var-mode} and the last row to the {\textit mixed-var distribution} setting. FI refers to the Farthest Insertion baseline\cite{rosenkrantz_analysis_2009}.}
\end{figure}

\begin{figure}[h!]
\centering
\hspace{-0.05in}\includegraphics[scale=0.52]{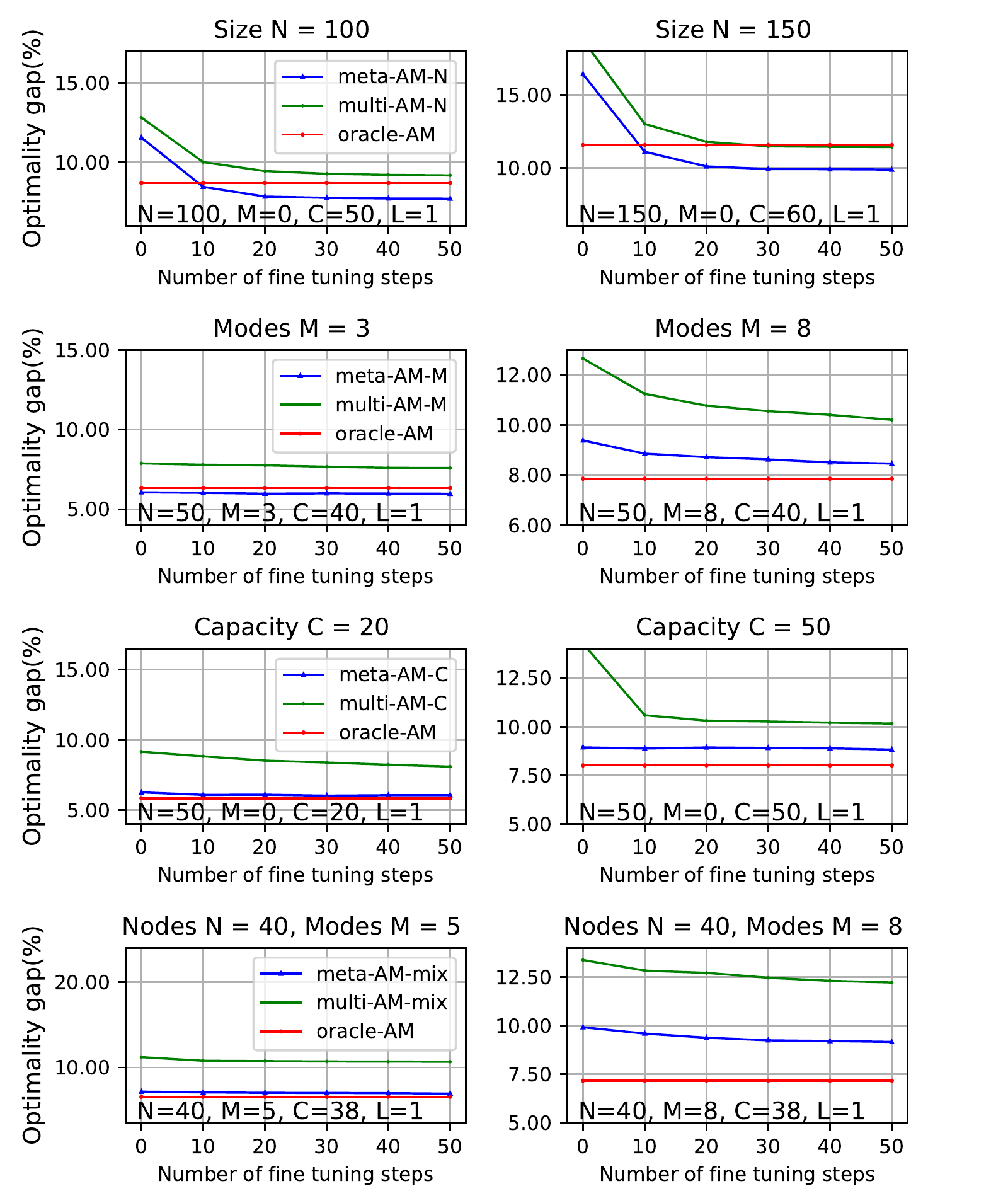}
\caption{\label{fig:meta-vs-multi_cvrp}
Test performance (optimality gap, in vertical axis) at different steps of fine-tuning (horizontal axis) of {\tt meta-AM} and {\tt multi-AM} models fine-tuned and tested on the CVRP distribution specified in the bottom left corner of each plot. The first row corresponds to {\textit var-size} setting, second  one to {\textit var-mode}, third row to the {\textit var-capacity} and the last row to the {\textit mixed-var distribution} setting.}
\end{figure}

\begin{figure}[h!]
\centering
\hspace{-0.15in}\includegraphics[scale=0.5]{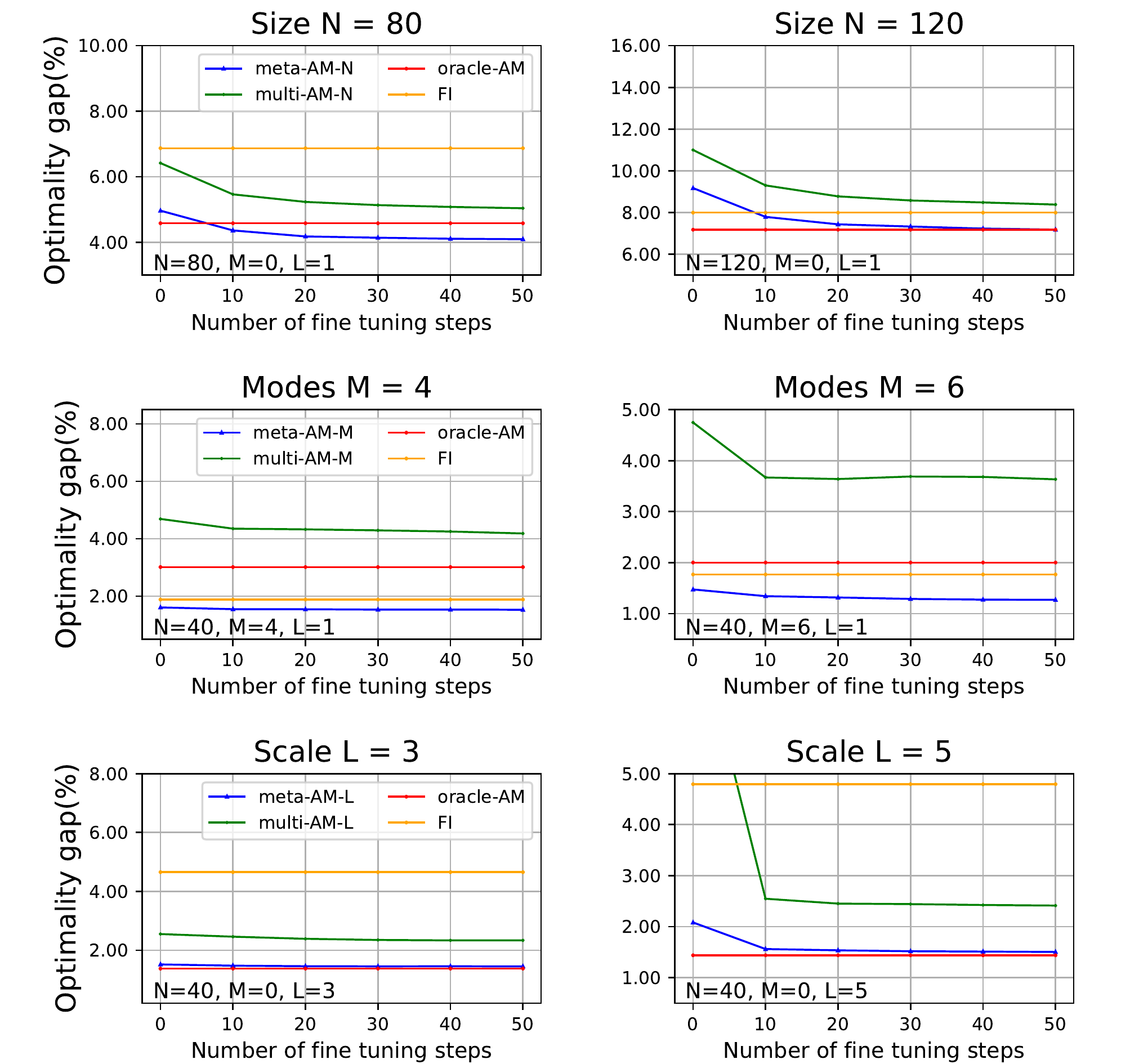}
\caption{\label{fig:meta-vs-multi_tsp_supp}  Test performance (average optimality gap, in vertical axis) at different steps of fine-tuning (horizontal axis) of {\tt meta-AM} and {\tt multi-AM}  models on the TSP distribution specified by the parameters at the bottom left corner of each plot. The first row corresponds to {\textit var-size} setting, second  one to the {\textit var-mode} and the last row to the {\textit var-scale} setting. FI refers to the Farthest Insertion baseline.}
\end{figure}

\begin{figure}[h!]
\centering
\hspace{-0.05in}\includegraphics[scale=0.55]{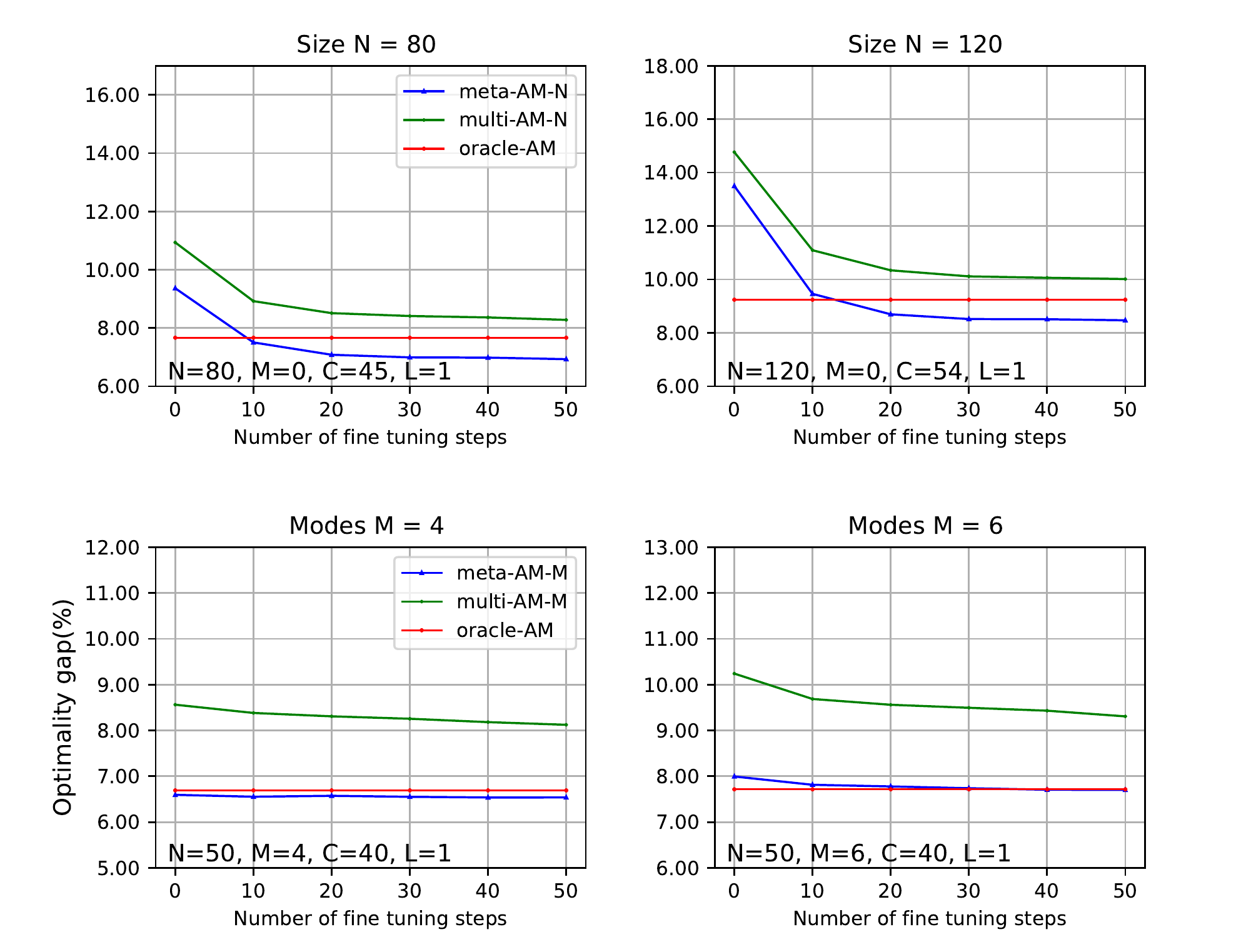}
\caption{\label{fig:meta-vs-multi_cvrp_supp}
Test performance (optimality gap, in vertical axis) at different steps of fine-tuning (horizontal axis) of {\tt meta-AM} and {\tt multi-AM} models fine-tuned and tested on the CVRP distribution specified in the bottom left corner of each plot. The first row corresponds to {\textit var-size} setting and the second  one to {\textit var-mode}.}
\end{figure}

\textbf{meta-AM Generalization: Evolution of performance at different fine-tuning steps:}
In the main paper, we showed the results for different datasets at  fine-tuning steps $K{=}0$ and $K{=}50$. In this section, in Fig.  ~\ref{fig:meta-vs-multi_tsp} and ~\ref{fig:meta-vs-multi_cvrp} we show the evolution of the optimality gaps at different fine-tuning steps varying from $K{=}0$ to $K{=}50$. Apart from the test datasets shown in main paper,  we also present additional results ({\tt meta-AM}) on additional tasks not shown in main paper in Fig.~\ref{fig:meta-vs-multi_tsp_supp} and ~\ref{fig:meta-vs-multi_cvrp_supp}.
Similar to the observations made in the main paper, we see that in most cases {\tt meta-AM} (blue) clearly generalizes better than the baseline {\tt multi-AM} (green) and even outperforms the {\tt oracle-AM} model. Similar observations can be made for  {\tt meta-AM} model for CVRP in Figure~\ref{fig:meta-vs-multi_cvrp_supp}. In all cases,  {\tt meta-AM} outperforms both  {\tt multi-AM} and  {\tt oracle-AM}.

\begin{figure}[h!]
\centering
\hspace{0in}\includegraphics[scale=0.5]{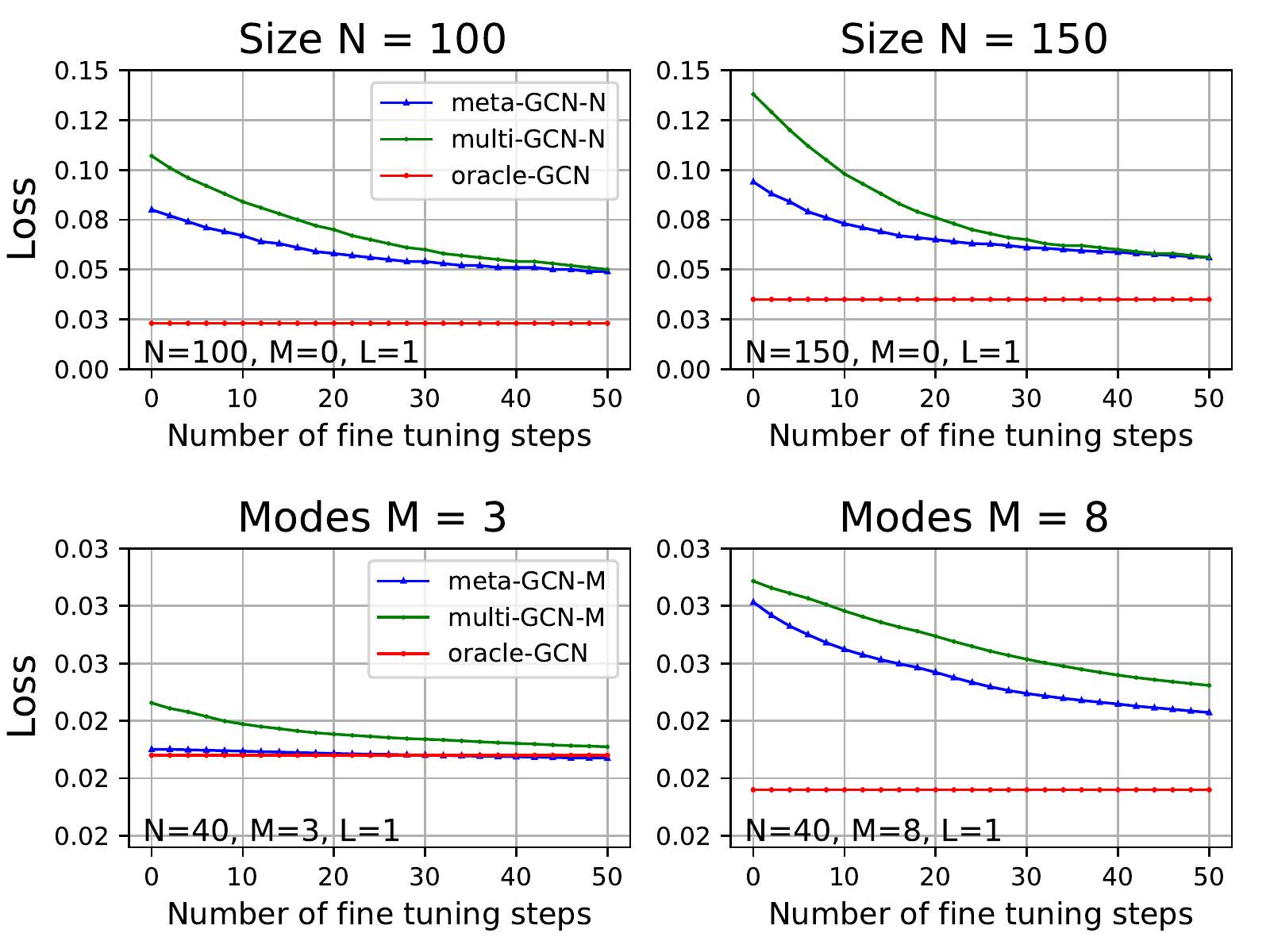}
\caption{\label{fig:meta-vs-multi_gcn_tsp} Performance (Test Loss, in vertical axis) at different steps of fine-tuning(horizontal axis) of {\tt meta-GCN} and {\tt multi-GCN} models fine-tuned and tested on TSP distribution specified in the bottom left corner of each plot. The first row corresponds to {\textit var-size} setting, the second to {\textit var-mode} setting. }
\label{app:testloss-tsp-gcn}
\end{figure}
\begin{figure}[h]
\includegraphics[scale=0.55]{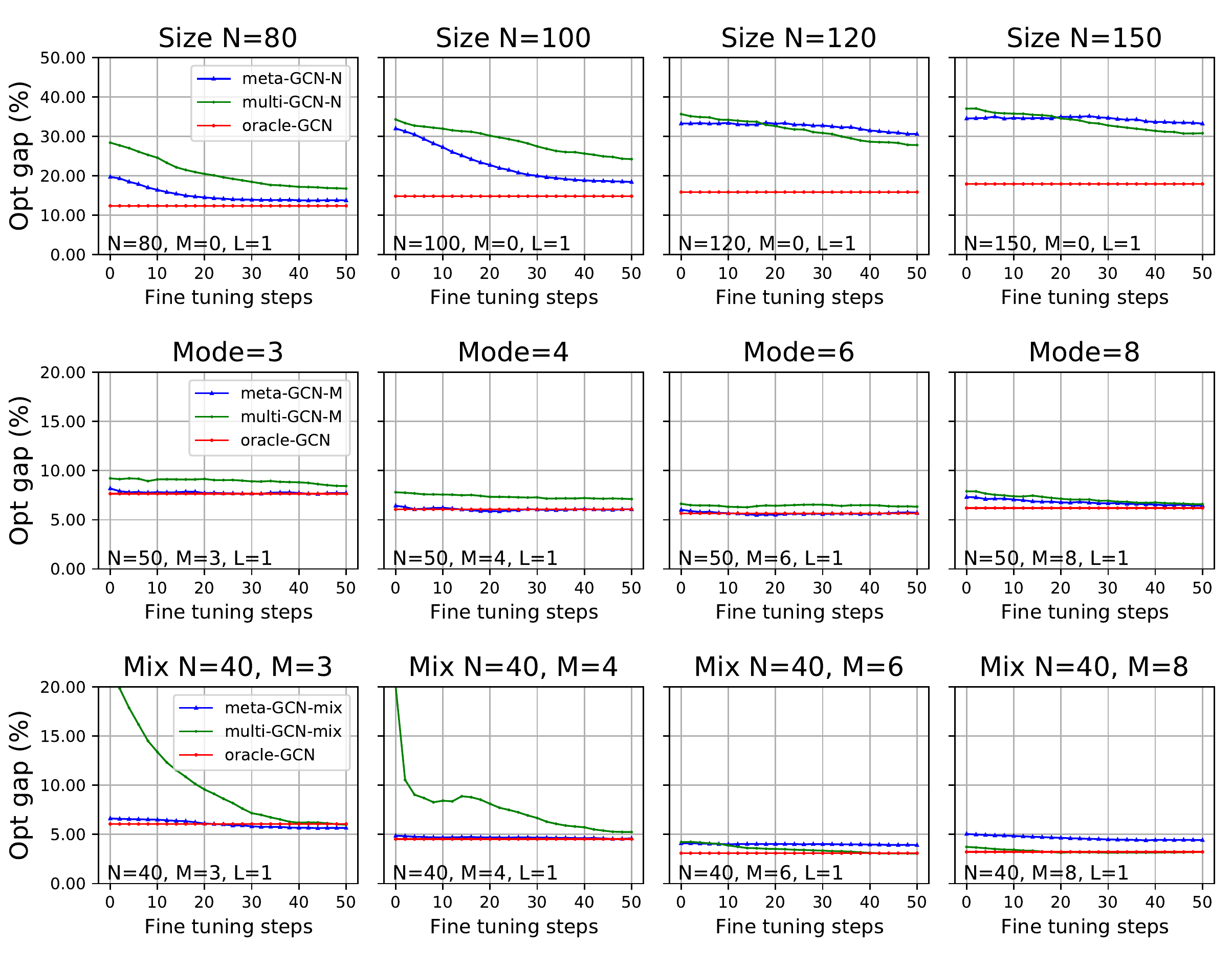}
\caption{\label{fig:meta-vs-multi_tsp_gcn_supp} Test performance (average optimality gap, in vertical axis) at different steps of fine-tuning (horizontal axis) of {\tt meta-GCN} and {\tt multi-GCN} on the TSP distribution specified by the parameters at the bottom left corner of each plot. The first row corresponds to {\textit var-size} setting, second  one to {\textit var-mode} and the last row to the {\textit mixed-var distribution} setting.
}
\end{figure}


\noindent \\
\textbf{meta-GCN Generalization: }
In Sec.5 of the main paper, we have presented results of the \texttt{meta-GCN} model on TSP. In this section, we show the evolution of performance of the \texttt{meta-GCN} model in terms of optimality gap.
In Fig.~\ref{fig:meta-vs-multi_tsp_gcn_supp} for TSP, for the \textit{var-mode setting} for \texttt{meta-GCN} trained on  $\task{N=50,M\in\{1,2,5\},L=1}$  we observe that in all cases, \texttt{meta-GCN} performs better than the baseline \texttt{multi-GCN} and in 3 out of 4 cases, it even outperforms the \texttt{oracle-GCN} model. For the \textit{var-size setting}, for \texttt{meta-GCN} trained on $\task{N\in\{10,20,30,50\},M=0,L=1}$, we observe that \texttt{meta-GCN} outperforms \texttt{multi-GCN} when the number of fine-tuning steps are less. However, we can see that when the number of fine-tuning steps increases, the optimality gap of \texttt{multi-GCN} is better even though its test loss is higher(see Fig.~\ref{app:testloss-tsp-gcn}). A possible reason for this behaviour is our use of greedy decoding and instead of the beam-search/sampling-based decoding. For the \textit{mixed-var distribution} setting, for the \texttt{meta-GCN} model trained on  {\scriptsize $(N,M) \in \{20,30,50\} \times \{1,2,5\}$}, we observe that for $(N{=}40,M{=}3)$ and $(N{=}40,M{=}4)$, \texttt{meta-GCN} outperforms \texttt{multi-GCN} and the \texttt{oracle-GCN} model in terms of optimality gap.

\begin{figure}[h!]
\centering
\hspace{0in}\includegraphics[scale=0.55]{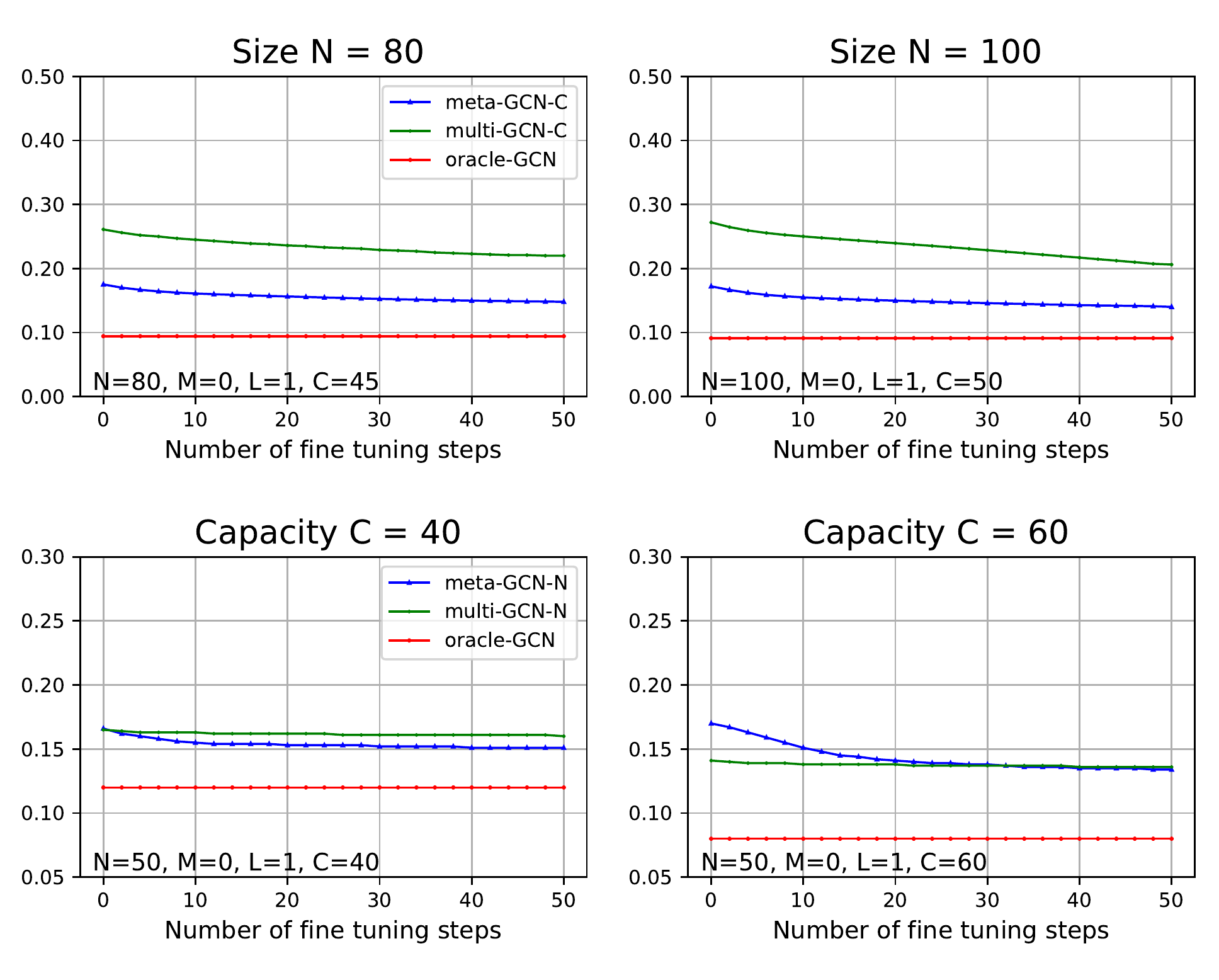}{-0.1in}
\caption{\label{fig:meta-vs-multi_gcn_vrp} Performance (Test Loss) at different steps of fine-tuning of {\tt meta-GCN} and {\tt multi-GCN} models fine-tuned and tested on CVRP distributions specified in the bottom left corner. The experiments are based upon the {\textit var-size} and {\textit var-capacity} settings. For training the \texttt{meta-GCN-C} and \texttt{multi-GCN-C} we use training instances with $\task{N={\in}\{20,30,50\},M=0,L=1,C{\in}\{30,35,40\}}$ (for {\textit var-size}) and $\task{N=50,M=0,L=1}$ and {\scriptsize $C{\in}\{20,30,50\}$}; (for {\textit var-capacity})}
\label{app:testloss-cvrp-gcn}
\end{figure}

\subsection{TSPlib: Instance wise results}
\label{TSPLIB:expts}
In this section we present the results for all the TSPlib instances used in Sec.~\ref{sec:realistic-datasets} of the main paper. In Table.~\ref{TSPLIB:tabl} we observe that in most cases, especially the larger size instances, our proposed meta-AM framework outperforms other baselines by a significant margin.
\begin{table}
\centering
\begin{tabular}{|l|c|c|c|}

\textbf{Instance} & AM($K{=}100$) & multi-AM($K{=}100$) & meta-AM($K{=}100$) \\ \hline
eil51 & \textbf{2.465} & 6.459 & 2.858 \\ \hline
berlin52 & 16.643 & 13.65 & \textbf{2.942} \\ \hline
st70 & \textbf{2.621} & 7.241 & 3.729 \\ \hline
eil76 & \textbf{5.233} & 7.688 & 6.202 \\ \hline
pr76 & \textbf{1.921} & 7.732 & 2.417 \\ \hline
rat99 & 14.224 & 25.941 & \textbf{13.511} \\ \hline
kroA100 & 13.413 & 17.705 & \textbf{5.08} \\ \hline
kroE100 & 9.157 & 9.557 & \textbf{6.963} \\ \hline
rd100 & 5.718 & 6.253 & \textbf{1.139} \\ \hline
kroC100 & \textbf{8.232} & 14.928 & 10.928 \\ \hline
kroB100 & 10.718 & 14.133 & \textbf{9.199} \\ \hline
kroD100 & 11.925 & 12.152 & \textbf{6.412} \\ \hline
eil101 & 5.862 & 9.291 & \textbf{5.228} \\ \hline
lin105 & 6.538 & 19.71 & \textbf{5.248} \\ \hline
pr107 & \textbf{5.348} & 8.698 & 7.353 \\ \hline
pr124 & 4.724 & 8.736 & \textbf{0.927} \\ \hline
bier127 & 12.281 & 23.628 & \textbf{7.312} \\ \hline
ch130 & 5.376 & 12.087 & \textbf{2.068} \\ \hline
pr144 & 11.163 & 14.412 & \textbf{5.909} \\ \hline
ch150 & \textbf{7.978} & 9.107 & 8.268 \\ \hline
kroA150 & 9.664 & 14.808 & \textbf{8.802} \\ \hline
kroB150 & 10.771 & 12.75 & \textbf{7.96} \\ \hline
pr152 & 10.201 & 9.848 & \textbf{5.038} \\ \hline
u159 & 10.95 & 21.465 & \textbf{7.317} \\ \hline
rat195 & \textbf{16.617} & 30.093 & 17.059 \\ \hline
d198 & 42.953 & 54.282 & \textbf{23.314} \\ \hline
kroA200 &\textbf{10.398} & 19.773 & 13.87 \\ \hline
kroB200 & 12.966 & 20.761 & \textbf{12.734} \\ \hline

\end{tabular}

\caption{ \label{TSPLIB:tabl}\textbf{TSPLib: Instance-wise results.} Comparison of optimality gaps achieved by AM, meta-AM and multi-AM models on TSPlib dataset after 100 steps of fine-tuning.}
\end{table}

\subsection{Visualisation of some solutions}
\begin{figure}[ht!]
\centering
\hspace{-0.15in}\includegraphics[scale=0.8]{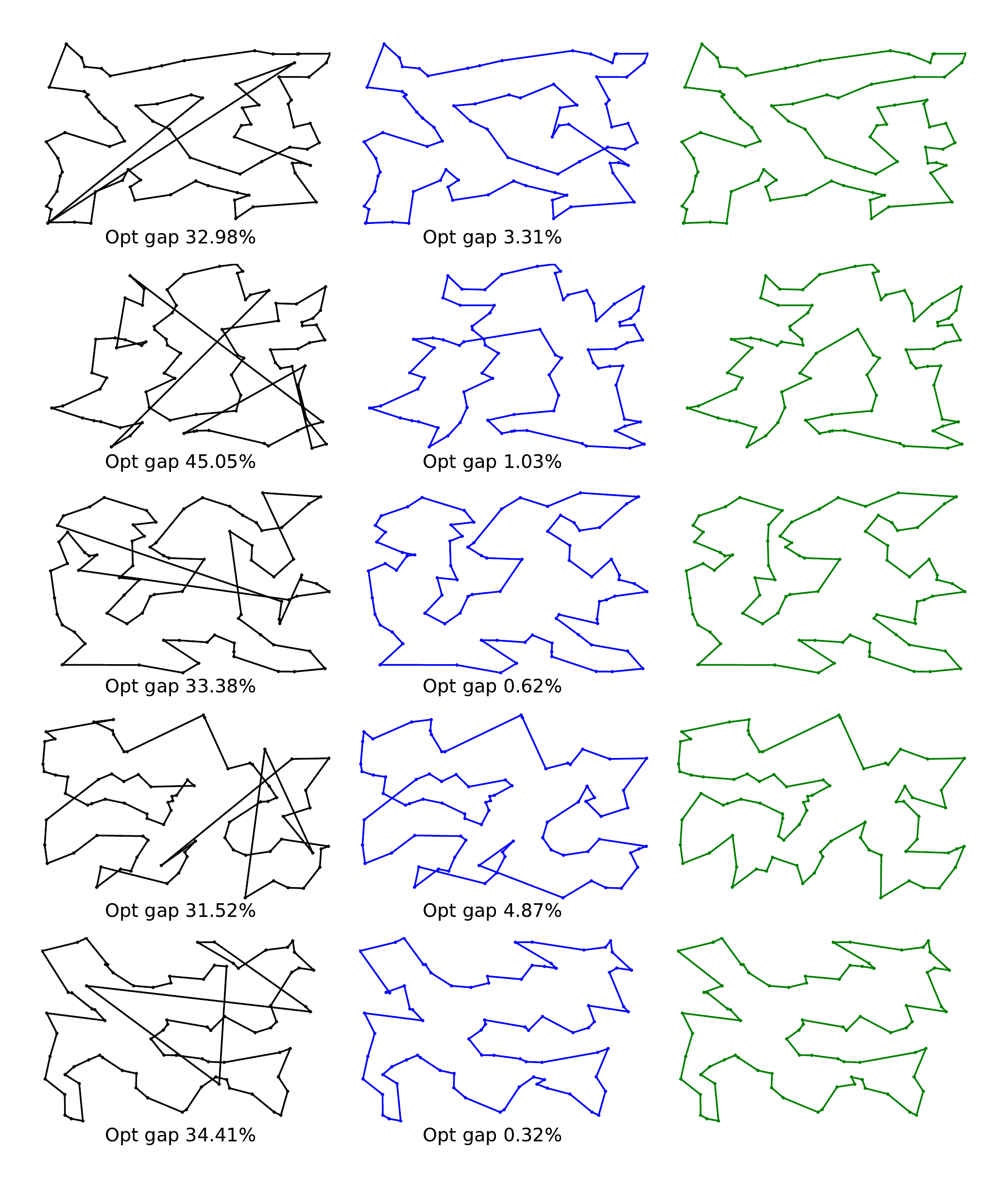}
\caption{\label{visualise:GCN} Solutions on some test graphs from $N{=}80,M=0,L=1$. Black: \texttt{meta-GCN-N} without fine-tuning; Blue: \texttt{meta-GCN-N} after 50 fine-tuning steps; Green: optimal solution}
\end{figure}

Figure~\ref{visualise:GCN} shows the solutions computed by 3 models on a few TSP instances sampled from $\mathcal{T}_{N=80,M=0,L=1}$, one per row. The first column (black) corresponds to \texttt{meta-AM} with no fine-tuning, trained for 24h on 1 million instances from $\mathcal{T}_{N\in\{10,20,30,50\},M=0,L=1}$ (hence not including the target distribution). The second column (blue) corresponds to the same model, after $50$ steps of fine tuning on $500$ instances drawn from the target distribution. The last column (green) corresponds to the true optimal results as obtained by Concorde.

\end{document}